\documentclass[runningheads]{llncs}
\usepackage{accv}

\usepackage{accvabbrv}
\usepackage{graphicx}
\usepackage{booktabs}
\usepackage[accsupp]{axessibility}  

\newcommand{\corrauth}{\unskip$^\star$}

\usepackage[breaklinks,colorlinks,citecolor=accvblue]{hyperref}
\usepackage{orcidlink}

\newcommand{\ours}{\textbf{{EvFRA}\xspace}}
\usepackage{multirow,enumitem,caption,array,makecell}
\usepackage{placeins}
\usepackage{dblfloatfix}
\usepackage[table]{xcolor}

\begin{document}
\title{Event-based Scene Synthesis via Inter-Frame Residual Alignment} 
\titlerunning{Event-based Scene Synthesis via Inter-Frame Residual Alignment}

\author{
Jiyun Kong\inst{1}\orcidlink{0009-0008-2259-9540} \and
Jun-Hyuk Kim\inst{2}\thanks{Corresponding authors}\orcidlink{0000-0002-7980-4583} \and
Jong-Seok Lee\inst{3,4}\corrauth\orcidlink{0000-0002-8038-1119}
}

\authorrunning{J. Kong et al.}

\institute{
Department of Artificial Intelligence, Yonsei University,
Republic of Korea\\
\and
Department of Artificial Intelligence, Chung-Ang University,
Republic of Korea\\
\and
School of Integrated Technology, Yonsei University,
Republic of Korea\\
\and
BK21 Graduate Program in Intelligent Semiconductor Technology, Yonsei University,
Republic of Korea\\
\email{\{jiyun.kong, jong-seok.lee\}@yonsei.ac.kr} \\
\email{junhyukkim@cau.ac.kr}
}

\maketitle

\begin{abstract}
    Event-based scene synthesis reconstructs target RGB frames from sparse image observations and asynchronous event streams, encompassing both video frame prediction and interpolation.
    Existing event-based synthesis methods commonly estimate optical flow to warp the observed frames toward the target time, but are vulnerable to inaccurate flow under large motion and occlusion and often rely on flow supervision or pretrained estimators.
    In this work, we propose \ours, an \textbf{Ev}ent-based scene synthesis framework based on inter-\textbf{F}rame \textbf{R}esidual \textbf{A}lignment.
    We identify a structural correspondence between event measurements and frame-to-frame scene changes, and exploit this correspondence for target frame synthesis.
    Our training pipeline consists of two stages: 1) an Event-to-Residual Alignment Variational Autoencoder (ER-VAE) aligns the event frame captured between the anchor and target frames with the corresponding inter-frame residual, and 2) a ControlNet-conditioned diffusion model is fine-tuned to denoise the residual latent using event data.
    Our method outperforms state-of-the-art methods by up to 2.61 dB and 1.85 dB in PSNR for frame prediction and interpolation, respectively, with consistent SSIM improvements.
    Code is available at \url{https://github.com/jiyun-kong/EvFRA}.

  \end{abstract}
\begin{figure}[t]
    \centering
    \begin{tabular}{@{}cccc@{}}
        \includegraphics[width=0.24\columnwidth]{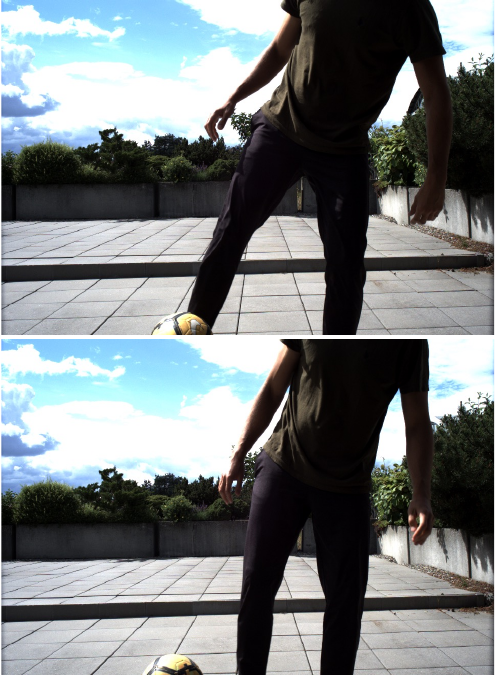} &
        \includegraphics[width=0.24\columnwidth]{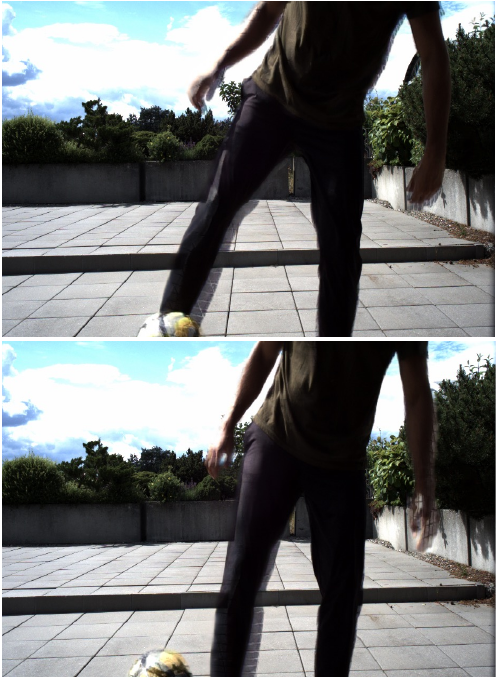} &
        \includegraphics[width=0.24\columnwidth]{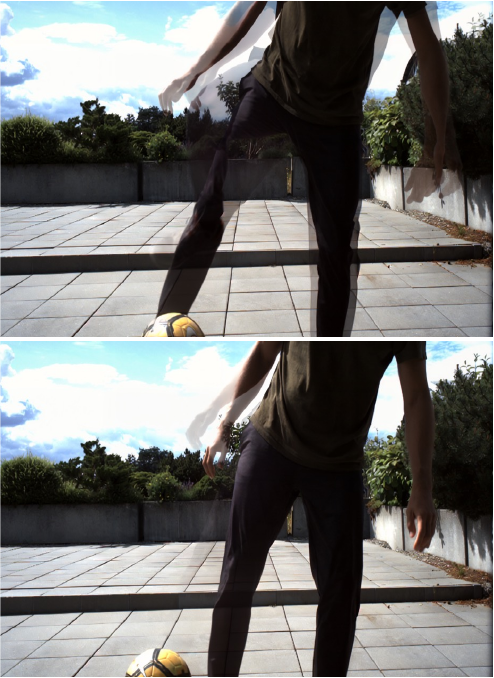} &
        \includegraphics[width=0.24\columnwidth]{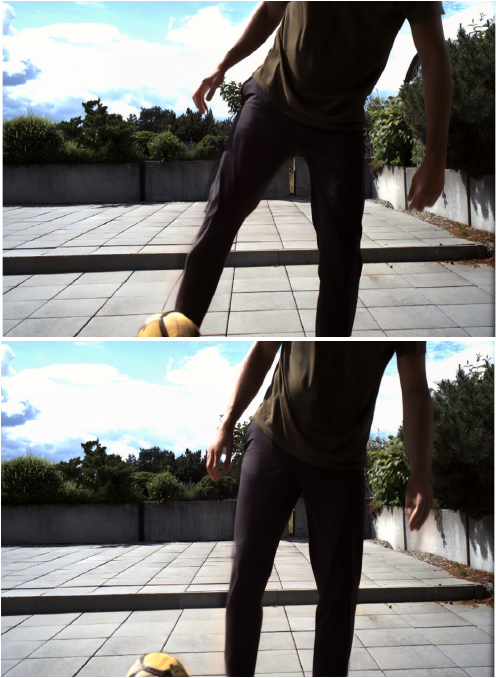} \\
        \small (a) Ground Truth &
        \small (b) VFPSIE &
        \small (c) CBMNet-Large &
        \small (d) \ours
    \end{tabular}
    \caption{Comparison of our method with other approaches on BS-ERGB~\cite{tulyakov2022timelenspp}, including VFPSIE~\cite{zhu2024video} and CBMNet-Large~\cite{kim2023cbmnet}. Ours produces more accurate motion prediction under fast movement, preserving object boundaries without ghosting artifacts.}
    \label{fig:teaser}
    \vspace{-6pt}
\end{figure}

\section{Introduction}
\label{sec:intro}
Video frame synthesis aims to generate unobserved frames within an existing video sequence, either between observed frames through interpolation or beyond them through frame prediction~\cite{liu2017voxel,hu2023dmvfn,zhang2024extdm}.
It is widely used in practical applications such as frame-rate conversion and slow-motion video generation~\cite{gao2022superfast,tulyakov2022timelenspp,kim2023cbmnet}.
However, accurately modeling rapid object motion, deformation, illumination changes, and camera movement from sparse RGB observations alone remains challenging~\cite{jiang2018super,bao2019depth,reda2022film}.
Event cameras~\cite{lichtsteiner2008dvs,gallego2020event,chakravarthi2024recent,ren2026systematic} provide complementary temporal cues by asynchronously recording per-pixel brightness changes between RGB frames.
With low sensing latency, a dynamic range exceeding 120 dB, and microsecond-level temporal resolution, they capture fine-grained scene dynamics under fast motion and challenging illumination, making them well suited for accurate video frame synthesis.

Existing event-based video frame synthesis methods typically use event data to estimate optical flow and propagate observed image content toward the target frame~\cite{bardow2016simultaneous,gehrig2021eraft,wan2022learning,gehrig2024dense}.
However, they often depend on optical flow supervision obtained from pretrained estimators~\cite{shiba2024secrets,barhaim2020scopeflow,jeong2024ocai}, whose errors may propagate to the synthesis model.
Moreover, optical flow-based warping itself frequently produces artifacts~\cite{liu2017voxel,liang2024flowvid} such as holes and blurring under large motion and occlusion, requiring additional inpainting or refinement modules~\cite{liu2017voxel,liang2024flowvid,zhu2024video,wang2025event,ni2023conditional}.

To address these limitations, we propose \ours, an event-based scene synthesis framework based on inter-frame residual alignment.
Instead of directly synthesizing the target frame, our method models the residual latent—the difference between the anchor and target frame latents—allowing the model to focus on inter-frame changes and enhance temporal consistency.
As shown in Figure~\ref{fig:teaser}, our method reconstructs fast motion better and preserves object boundaries with fewer ghosting artifacts compared to existing approaches.
Our framework further builds on a diffusion model pretrained on large-scale image datasets~\cite{sohl2015deep,ho2020denoising,dhariwal2021diffusion,rombach2022high,blattmann2023videoldm}, leveraging its generative prior while conditioning the denoising process on event data to control temporal dynamics.

The training process of our method consists of two stages.
In stage 1, a VAE-based model named Event-to-Residual Alignment Variational Autoencoder (ER-VAE) is trained to align the event frame with the corresponding residual latent.
By fine-tuning the VAE encoder from the pretrained Stable Diffusion model, the event latent is encouraged to follow a distribution consistent with that of the residual latent.
In stage 2, a conditional diffusion model is trained to denoise the residual latent under image and event conditioning.
Unlike diffusion models trained in the high-dimensional pixel space~\cite{sohl2015deep,ho2020denoising,ho2022cascaded,peebles2023scalable}, our approach performs generation in a low-dimensional residual latent space that focuses on frame differences.
During inference, where the target frame is unobserved, the event latent from the trained ER-VAE serves as an initialization, enabling the diffusion model to synthesize the target frame.
Our method is flexible so that it can be applied to both video frame interpolation and video frame prediction.

Our contributions are summarized as follows: 
\vspace{-0.5em}
\begin{itemize}[leftmargin=*,label=\textbf{·}]
    \item \textbf{Inter-Frame Residual Alignment:}
    We introduce ER-VAE to explicitly align event representations with inter-frame residual latents. 
    Rather than using events as auxiliary conditioning signals, ER-VAE learns the event latent as a proxy for the anchor-to-target latent displacement. 
    At inference time, this event-derived residual representation provides an informed initialization for residual denoising.
    \item \textbf{Residual-Space Scene Synthesis:}
    We propose a two-stage framework consisting of ER-VAE and an event-conditioned latent diffusion model. 
    By generating the residual between observed and target frames, our formulation explicitly models inter-frame changes and promotes stable motion alignment across consecutive frames. 
    To the best of our knowledge, ours is the first event-based scene synthesis framework to model events as proxies for inter-frame residuals.
    \item \textbf{Variable-Length Temporal Augmentation:} 
    We introduce Variable-Length Temporal (VLT) Augmentation, which progressively increases the temporal intervals used during training to expose the model to a broader range of motion patterns. This strategy improves generalization across varying motion magnitudes and synthesis intervals while maintaining stable optimization from short to longer temporal ranges.
\end{itemize}

\section{Related Work}
\label{sec:related}

\subsection{Event-based Frame Synthesis}
Event cameras asynchronously record brightness changes, providing motion-sensitive observations that benefit frame synthesis in dynamic scenes~\cite{lichtsteiner2008dvs,gallego2020event,chakravarthi2024recent}.
Research on event-based frame synthesis can be categorized into three settings. First, event-to-frame reconstruction methods recover intensity images directly from event streams, as in E2VID~\cite{rebecq2019events}, HyperE2VID~\cite{ercan2024hypere2vid}, and E2HQV~\cite{qu2024e2hqv}. 
Second, event-based frame interpolation synthesizes intermediate frames from surrounding RGB frames and events, as in TimeLens~\cite{tulyakov2021time}, TimeLens++~\cite{tulyakov2022timelenspp}, SuperFast~\cite{gao2022superfast}, and CBMNet~\cite{kim2023cbmnet}.
Third, event-based frame prediction extrapolates a target frame from an anchor RGB frame and subsequent events, as demonstrated by VFPSIE~\cite{zhu2024video}.

Many event-based frame synthesis methods rely on optical flow to establish motion correspondences and propagate observed image content toward the target frame~\cite{bardow2016simultaneous,gehrig2021eraft,wan2022learning,gehrig2024dense}. 
Although some methods incorporate additional synthesis or refinement modules, their reconstructions remain strongly dependent on the estimated motion fields~\cite{tulyakov2022timelenspp,kim2023cbmnet,zhu2024video,wang2025event}. 
In practice, optical-flow supervision is often obtained from pretrained flow estimators (e.g., E-RAFT~\cite{gehrig2021eraft}), whose estimation errors may propagate to the subsequent synthesis model~\cite{shiba2024secrets,barhaim2020scopeflow,jeong2024ocai}. 
Moreover, warping itself is limited in its ability to recover content in occluded or newly revealed regions~\cite{liu2017voxel,liang2024flowvid,jeong2024ocai}. 
In contrast, our method avoids explicit optical flow estimation and flow-based pixel warping by directly modeling inter-frame residuals aligned with event representations.

\subsection{Conditional Diffusion Models}
Diffusion models have recently emerged as a dominant paradigm in image generative modeling~\cite{sohl2015deep,ho2020denoising,dhariwal2021diffusion}, based on a reverse diffusion process that employs a U-Net-based denoising network to progressively remove Gaussian noise.
Latent diffusion models further improve computational efficiency and visual fidelity by performing diffusion in a compact latent feature space, where high-dimensional images are encoded using an autoencoder such as a VAE~\cite{kingma2013auto,rombach2022high}.
This latent-space formulation enables efficient high-resolution generation and flexible conditioning, facilitating the extension of diffusion models to diverse vision tasks and conditioning modalities such as text, sketches, and poses~\cite{rombach2022high,zhang2023adding,blattmann2023videoldm,chen2023controlavideo,saxena2024surprising,sun2026transient,kai2026evtexturepp}.

Recent studies have further explored diffusion models for event-conditioned frame synthesis.
E-Motion~\cite{wu2024emotion} employs pretrained Stable Video Diffusion together with reinforcement learning to generate future event sequences.
For event-based frame interpolation, EGVD~\cite{zhang2025egvd} and RE-VDM~\cite{chen2025repurposing} adapt pretrained Stable Video Diffusion by incorporating event information as motion guidance.
EventDiff~\cite{zheng2025eventdiff} introduces an event-frame hybrid autoencoder to fuse event streams with RGB frames through a latent-space denoising diffusion process.
In contrast, our method builds on pretrained Stable Diffusion and performs event-guided generation in the inter-frame residual latent space, allowing the diffusion process to focus directly on frame-to-frame changes.

\begin{figure}[t]
    \centering
    \includegraphics[width=0.90\columnwidth]{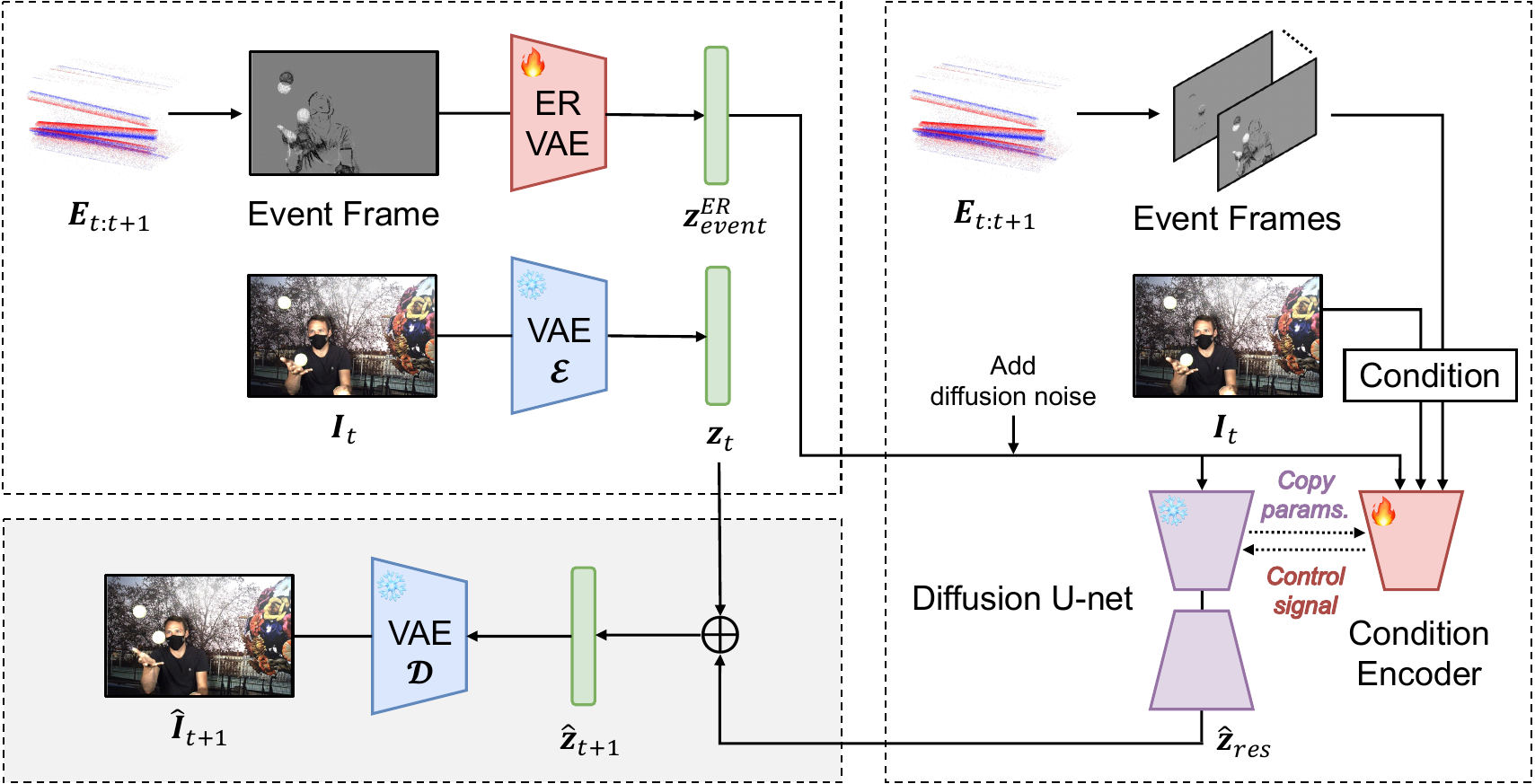}
    \caption{Overview of our proposed method \ours. During inference, Gaussian noise at $\sigma_{\max}$ is added to $z^\text{ER}_\text{event}$ to initialize the denoising trajectory.
    The diffusion U-net predicts the residual latent $\hat{z}_\text{res}$, which is added to $z_t$ to obtain the predicted latent $\hat{z}_{t+1}$.
    Finally, the decoder reconstructs the target frame $\hat{I}_{t+1}$ from $\hat{z}_{t+1}$.}
    \label{fig:overview}
    \vspace{-6pt}
\end{figure}

\begin{figure*}[t]
    \centering
    \includegraphics[width=0.98\textwidth]{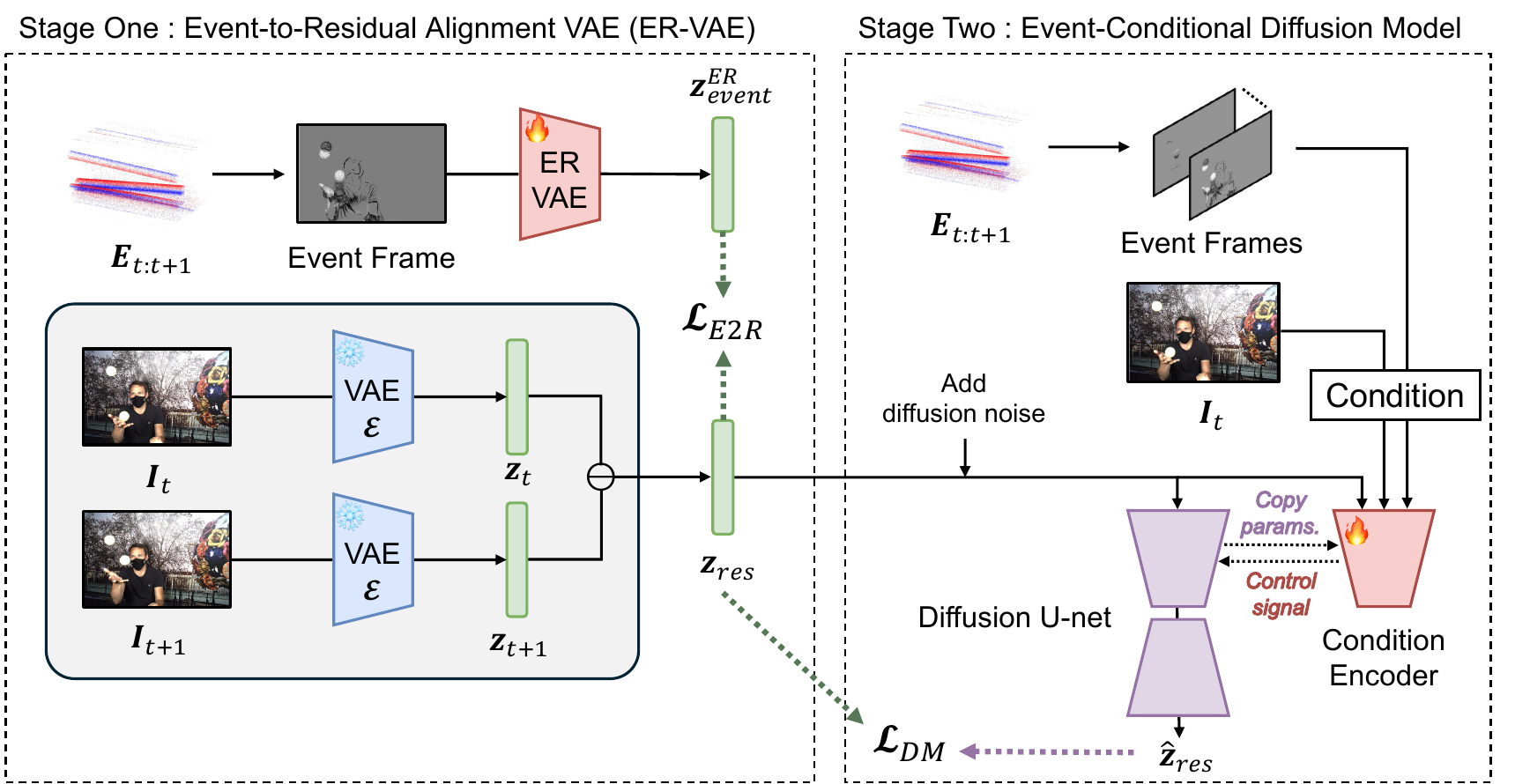}
    \caption{Training pipeline of \ours. The framework consists of two stages: in stage 1, ER-VAE is optimized with the loss $\mathcal{L}_\text{E2R}$, which jointly enforces latent alignment between the event latent $z^\text{ER}_\text{event}$ and the residual latent $z_\text{res} = z_{t+1} - z_t$.
    In stage 2, the event-conditioned encoder is trained with the denoising objective $\mathcal{L}_\text{DM}$ to guide the diffusion model in predicting the denoised residual latent $\hat{z}_\text{res}$.}
    \label{fig:training}
    \vspace{-6pt}
  \end{figure*}
  
\section{Proposed Method}
\label{sec:method}
\subsection{Framework Overview}
    An overview of our proposed method, \ours, is illustrated in Figure~\ref{fig:overview}.
    Given an anchor frame $I_t$ and the corresponding event stream $E_{t:t+1}$, the Event-to-Residual Alignment VAE (ER-VAE) encodes the event frame into a latent representation $z^\mathrm{ER}_\mathrm{event}$, which is aligned with the inter-frame residual latent.
    Meanwhile, the pretrained Stable Diffusion VAE encoder extracts the anchor latent $z_t$ from $I_t$.
    During inference, Gaussian noise at $\sigma_{\max}$ is added to $z^\mathrm{ER}_\mathrm{event}$ to initialize the denoising trajectory.
    $I_t$ is encoded as image conditioning and provided through the cross-attention layers, while event frames from $E_{t:t+1}$ are provided to the ControlNet for guided generation.
    The predicted residual latent $\hat{z}_\mathrm{res}$ is added to $z_t$ to form the target latent $\hat{z}_{t+1}$, which is then decoded into the final predicted frame $\hat{I}_{t+1}$ using the VAE decoder.

    The training process consists of two stages (Figure~\ref{fig:training}): In stage 1, the ER-VAE is trained to align the event latent with the residual latent (Section~\ref{sec:stage1}).
    In stage 2, the Event Conditional Diffusion Model is trained to denoise the residual latent conditioned on the event information (Section~\ref{sec:stage2}). 
    The same anchor-to-target residual synthesis process is used for both frame prediction and interpolation, with task-specific formulations for each setting (Section~\ref{sec:prediction_and_interpolation}).
    In both stages, we apply the Variable-Length Temporal (VLT) Augmentation strategy to enhance temporal generalization and motion robustness (Section~\ref{sec:vlt}).

\begin{figure}[t]
    \centering
    \includegraphics[width=\columnwidth]{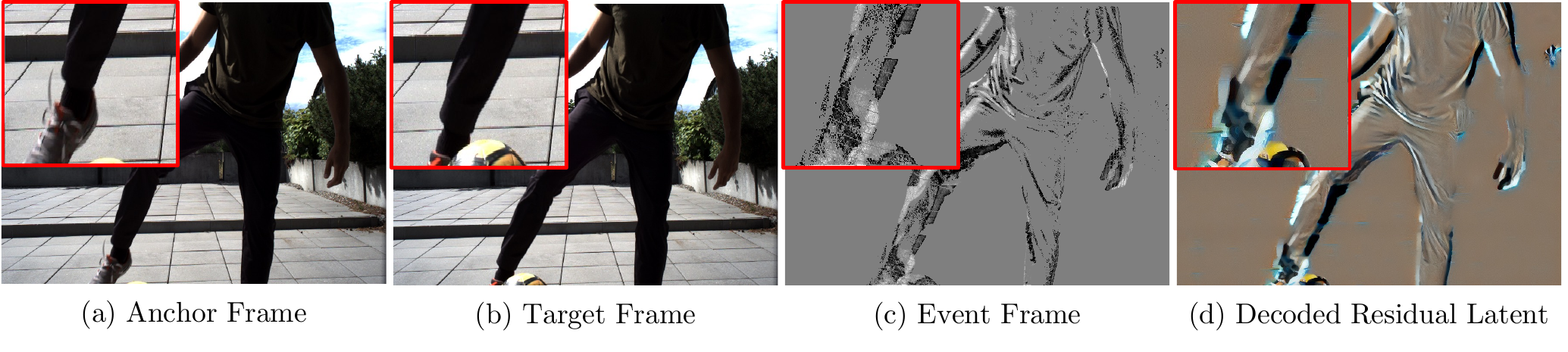}
    \caption{Structural relationship between event and residual representations.
    The decoded residual latent (d) is derived from the anchor (a) and target frame (b) as the difference between them. The event frame (c) reveals that event data captures inter-frame motion consistent with the decoded residual latent.}
    \label{fig:event_residual_relationship}
\end{figure}

\subsection{Stage One: Event-to-Residual Alignment VAE}
\label{sec:stage1}
The goal of the ER-VAE is to align the event representation with the inter-frame residual latent. 
The event frame is generated by accumulating the event stream $E_{t:t+1}$ captured between the anchor frame $I_t$ and the target frame $I_{t+1}$. 
It provides a compact encoding of temporal motion and brightness changes.
Meanwhile, the VAE encoder extracts the anchor and target latents, $z_t$ and $z_{t+1}$, from which we define the inter-frame residual latent as $z_\mathrm{res} = z_{t+1} - z_t$.
Therefore, the event latent $z^\mathrm{ER}_\mathrm{event}$ extracted from the event frame and the residual latent $z_\mathrm{res}$ capture the same motion information from different observation spaces, and their alignment facilitates a unified latent representation of motion dynamics. 

Figure~\ref{fig:event_residual_relationship} visually illustrates this relationship. 
Compared to the anchor frame (a), the target frame (b) exhibits motion, including the emergence of the ball and the displacement of the leg.
These dynamic changes are clearly reflected in the event frame (c), where motion boundaries around the moving leg and the newly appearing ball are prominently highlighted.
The decoded residual latent (d), derived from the difference between the anchor and target frames, exhibits structurally similar patterns to the event frame.
This visual correspondence demonstrates that the event representation effectively approximates the residual latent.

ER-VAE is based on Stable Diffusion's autoencoder, where the encoder and quantization module are fine-tuned. 
It receives the event frame as input and jointly optimizes two objectives: latent alignment and image reconstruction.
The combined loss is defined as follows:
\begin{equation}
\mathcal{L}_\mathrm{E2R} = \|\hat{z}_\mathrm{res} - z_\mathrm{res}\|_1 + \gamma \|\hat{I}_{t+1} - I_{t+1}\|^2_2,
\label{eq:er_vae_loss}
\end{equation}
where $\hat{I}_{t+1}$ denotes the reconstructed image obtained by decoding the predicted latent $\hat{z}_{t+1}$ through the VAE decoder, which is shifted in the latent space by the predicted residual $\hat{z}_\mathrm{res}$ from $z_t$.
The parameter $\gamma$ controls the weight of the image reconstruction loss. 
In our experiments, $\gamma$ is progressively increased from 0.3 to 0.7 using a cosine scheduling strategy, gradually emphasizing reconstruction quality.
Furthermore, instead of enforcing a KL divergence loss often adopted in conventional VAEs, ER-VAE learns a deterministic residual mapping by aligning latent means, which yields more consistent latent representations for the subsequent diffusion stage, consistent with the findings of Ghosh et al.~\cite{ghosh2019variational} (see Section~\ref{subsec:ablation}).

\subsection{Stage Two: Event-Conditional Diffusion Model}
\label{sec:stage2}
In stage 2, we train an event-conditioned diffusion model to denoise the inter-frame residual latent.
Based on Stable Diffusion's pretrained weights, the U-net is frozen, while only the ControlNet~\cite{zhang2023adding} and the conditional encoder are fine-tuned. 
ControlNet creates a trainable copy by copying the parameters of
existing diffusion model blocks and inserts the control signal into the diffusion feature. 
Zero-initialized convolution layers connect frozen blocks with learnable blocks, enabling the original diffusion model to retain its expressive power while acquiring external control capabilities through ControlNet training. 
The model receives two types of conditioning signals—image and event conditions—that provide spatial and motion information. The image condition combines semantic and spatial representations of the frame, obtained from the CLIP Vision
Encoder~\cite{radford2021learning} and Latent Tokenization. These representations are concatenated and fed into the cross-attention layers of the U-net. 
The event condition is provided to ControlNet as stacked event frames constructed with different temporal accumulation windows, enabling the model to capture motion patterns across multiple temporal scales.

The learning process follows the Elucidated Diffusion Model (EDM)~\cite{karras2022elucidating}.
During training, Gaussian noise with noise level $\sigma$, sampled from a log-normal distribution, is added to the residual latent $z_\mathrm{res}$.
For video frame prediction (VFP), the diffusion model is optimized with

\begingroup
\small
\begin{equation}
\begin{aligned}
\mathcal{L}_\mathrm{DM}^{\mathrm{VFP}}
=
\mathbb{E}_{z_\mathrm{res},\,\sigma}
\Big[
& w(\sigma)
\left\lVert
\hat{z}_\mathrm{res}
-
z_\mathrm{res}
\right\rVert_2^2 + \lambda_\mathrm{LPIPS}
\mathcal{L}_\mathrm{LPIPS}
(\hat{I}_{t+1}, I_{t+1}) \\
&+ \lambda_\mathrm{L1}
\left\lVert
\hat{I}_{t+1}
-
I_{t+1}\right\rVert_1 
+ \lambda_\mathrm{ev}
\mathcal{L}_\mathrm{ev\text{-}L1}
(\hat{I}_{t+1}, I_{t+1})
\Big].
\end{aligned}
\label{eq:VFP_DM}
\end{equation}
\endgroup

\noindent
Here, $w(\sigma)$ denotes the noise-dependent EDM weighting term.
The LPIPS term~\cite{zhang2018unreasonable} encourages perceptual similarity, while the $L_1$ term enforces pixel-level reconstruction fidelity.
The event-weighted loss $\mathcal{L}_\mathrm{ev\text{-}L1}$ emphasizes reconstruction errors in event-active regions by weighting the pixel-wise $L_1$ loss using a per-sample normalized event activity map.
We set $\lambda_\mathrm{LPIPS}=0.3$ and $\lambda_\mathrm{L1}=1.0$.
For the event-weighted term, $\lambda_\mathrm{ev}$ is set to $0.5$.

\subsection{Video Frame Prediction and Interpolation}
\label{sec:prediction_and_interpolation}
\ours\ supports both video frame prediction and video frame interpolation using the same event-to-residual alignment and residual denoising framework.
In both tasks, ER-VAE maps events over an anchor-to-target interval into the corresponding inter-frame residual latent space.
The primary difference lies in the number of residual branches supervised by the event-conditioned diffusion model and the reconstruction of the final target frame.

For video frame prediction, the aligned event latent
$z_\mathrm{event}^\mathrm{ER}$ initializes the residual denoising process for a single anchor-to-target branch. The model is trained according to the loss function in Eq.~\ref{eq:VFP_DM}.
For video frame interpolation, the target frame $I_t$ lies between two anchor
frames $I_{t_1}$ and $I_{t_2}$, where $t_1 < t < t_2$.
The same ER-VAE and event-conditioned diffusion model are applied to both
anchor-to-target directions.
ER-VAE produces the aligned event latents
$z_\mathrm{event}^\mathrm{fwd}$ and $z_\mathrm{event}^\mathrm{bwd}$
from the corresponding forward and backward event intervals, which initialize
the two residual denoising branches.
The VFI objective is defined as

\begingroup
\scriptsize
\begin{equation}
\begin{aligned}
\mathcal{L}_\mathrm{DM}^{\mathrm{VFI}}
=
\mathbb{E}_{z_\mathrm{res}^{\mathrm{fwd}},
            z_\mathrm{res}^{\mathrm{bwd}},\,\sigma}
\Big[
& w(\sigma)
\Big(
\left\lVert
\hat{z}_\mathrm{res}^{\mathrm{fwd}}
-
z_\mathrm{res}^{\mathrm{fwd}}
\right\rVert_2^2
+
\left\lVert
\hat{z}_\mathrm{res}^{\mathrm{bwd}}
-
z_\mathrm{res}^{\mathrm{bwd}}
\right\rVert_2^2
\Big)
\\
&+
\lambda_\mathrm{LPIPS}
\mathcal{L}_\mathrm{LPIPS}(\hat{I}_t,I_t)
+
\lambda_\mathrm{L1}
\left\lVert
\hat{I}_t-I_t
\right\rVert_1
+
\lambda_\mathrm{ev}
\mathcal{L}_\mathrm{ev\text{-}L1}(\hat{I}_t,I_t)
\Big].
\end{aligned}
\label{eq:VFI_DM}
\end{equation}
\endgroup

\noindent Compared to Eq.~\ref{eq:VFP_DM}, Eq.~\ref{eq:VFI_DM} supervises the forward and backward latent residual predictions, $\hat{z}_\mathrm{res}^\mathrm{fwd}$ and $\hat{z}_\mathrm{res}^\mathrm{bwd}$, against $z_\mathrm{res}^\mathrm{fwd}$ and $z_\mathrm{res}^\mathrm{bwd}$, respectively.
The forward and backward residual terms therefore jointly optimize the forward
and backward anchor-to-target directions and promote bidirectional consistency in the residual latent space. 
The blended prediction $\hat{I}_t$ used in all three image-space loss terms is defined as
\begingroup
\scriptsize
\begin{equation}
\hat{I}_t = w_{\mathrm{fwd}} \hat{I}_t^{\mathrm{fwd}} + w_{\mathrm{bwd}} \hat{I}_t^{\mathrm{bwd}}, \qquad w_{\mathrm{fwd}} = \frac{d_{\mathrm{bwd}}}{d_{\mathrm{fwd}} + d_{\mathrm{bwd}}}, \quad w_{\mathrm{bwd}} = \frac{d_{\mathrm{fwd}}}{d_{\mathrm{fwd}} + d_{\mathrm{bwd}}}.
\end{equation}
\endgroup
\noindent For VFI, the forward and backward event activity maps are merged by a pixel-wise maximum before computing the event-weighted $L_1$ loss on the blended prediction $\hat{I}_t$.
Here, $\hat{I}_t^{\mathrm{fwd}}$ and $\hat{I}_t^{\mathrm{bwd}}$ denote the forward and backward pixel-space predictions, respectively, while $d_{\mathrm{fwd}}$ and $d_{\mathrm{bwd}}$ represent their temporal distances to the target frame.
Accordingly, the prediction generated from the temporally closer anchor frame receives a larger fusion weight.
Together, these modifications extend the single-branch prediction formulation to interpolation through bidirectional residual supervision and temporally aware fusion, while retaining the same ER-VAE and event-conditioned diffusion architecture.

\subsection{Variable-Length Temporal Augmentation}
\label{sec:vlt}
During training, we introduce Variable-Length Temporal (VLT) Augmentation, a curriculum-style strategy~\cite{zhou2021curriculum} that progressively increases the temporal interval between the anchor and target frames.
The model first learns from short temporal intervals to ensure stable convergence and accurate reconstruction of small motion.
As training progresses, longer temporal intervals are introduced, encouraging the model to capture long-range dependencies and dynamic motion patterns more effectively.
This progressive schedule improves generalization across diverse motion magnitudes and synthesis intervals.
Ablation results analyzing the effectiveness of VLT Augmentation are presented in Section~\ref{subsec:ablation}.

\section{Experiments}
\label{sec:experiments}

\subsection{Datasets}
\label{subsec:datasets}
We use three event-based video datasets: BS-ERGB~\cite{tulyakov2022timelenspp}, HS-ERGB~\cite{tulyakov2021time}, and GoPro~\cite{nah2017deep}.
BS-ERGB contains event–RGB pairs captured at 28 fps with a resolution of $970\times625$.
Following~\cite{tulyakov2021time}, we evaluate both video frame prediction and interpolation under the 1-frame and 3-frame settings.
Here, ``$N$-frame setting'' means that in the case of prediction, the RGB frame at time $t+N$ is predicted given the frame at time $t$, and for interpolation, the frame at $t$ is synthesized using the frames at $t-N$ and $t+N$.
HS-ERGB consists of sequences with variable frame rates ($150$–$163$ fps) and diverse resolutions (e.g., $957\times856$, $430\times500$).
Following~\cite{tulyakov2021time}, we evaluate under the 7-frame setting.
GoPro is a synthetic dataset derived from 240 fps, $1280\times720$ RGB videos and converted into event data using the event simulator~\cite{rebecq2018esim}.
Following~\cite{zhu2024video}, we evaluate under the 7-frame and 15-frame settings.
Our model is trained on the BS-ERGB training set and evaluated on the BS-ERGB, HS-ERGB, and GoPro test sets.

\begin{table*}[t]
\centering
\caption{Quantitative comparison for video frame prediction and interpolation on the real-world event datasets: BS-ERGB~\cite{tulyakov2022timelenspp} and HS-ERGB~\cite{tulyakov2021time}. The best and second-best results are highlighted in \textbf{bold} and \underline{underline}, respectively.}
\label{tab:main_results_real}
\small
\renewcommand{\arraystretch}{1.15}
\resizebox{\textwidth}{!}{
\begin{tabular}{c|c|ccc|ccc|ccc}
\hline
\multirow{3}{*}{Task} & \multirow{3}{*}{Method} &
\multicolumn{6}{c|}{BS-ERGB~\cite{tulyakov2022timelenspp}} &
\multicolumn{3}{c}{HS-ERGB~\cite{tulyakov2021time}} \\
\cline{3-11}
& & \multicolumn{3}{c|}{1 frame} & \multicolumn{3}{c|}{3 frames} &
\multicolumn{3}{c}{7 frames} \\
\cline{3-11}
& & PSNR$\uparrow$ & SSIM$\uparrow$ & LPIPS$\downarrow$ &
PSNR$\uparrow$ & SSIM$\uparrow$ & LPIPS$\downarrow$ &
PSNR$\uparrow$ & SSIM$\uparrow$ & LPIPS$\downarrow$ \\
\hline
\hline
\multirow{2}{*}{\makecell[c]{Frame\\Prediction}}
& VFPSIE~\cite{zhu2024video} (AAAI'24)
& \underline{24.41} & \underline{0.805} & \textbf{0.091} & \underline{23.20} & 0.782 & \textbf{0.114} & \underline{28.81} & \underline{0.845} & \underline{0.112} \\
& \textbf{\ours (VFP)}
& \textbf{25.27} & \textbf{0.895} & \underline{0.100} & \textbf{24.20} & \textbf{0.804} & \underline{0.119} & \textbf{29.74} & \textbf{0.858} & \textbf{0.101} \\
\hline
\multirow{5}{*}{\makecell[c]{Frame\\Interpolation}}
& TimeLens~\cite{tulyakov2021time} (CVPR'21)
& 23.08 & 0.694 & 0.121 & 21.93 & 0.658 & 0.172 & 25.98 & 0.793 & 0.167 \\
& CBMNet-Large~\cite{kim2023cbmnet} (CVPR'23)
& \underline{24.96} & \underline{0.827} & 0.120 & \underline{23.83} & \underline{0.800} & \underline{0.123} & \underline{28.14} & \underline{0.844} & \underline{0.149} \\
& RE-VDM~\cite{chen2025repurposing} (CVPR'25)
& 24.88 & 0.725 & \underline{0.118} & 22.03 & 0.660 & 0.131 & 24.71 & 0.802 & 0.156 \\
& EGVD~\cite{zhang2025egvd} (ICCVW'25)
& 20.00 & 0.581 & 0.176 & 18.73 & 0.517 & 0.181 & 23.60 & 0.714 & 0.198 \\
& \textbf{\ours (VFI)}
& \textbf{26.81} & \textbf{0.899} & \textbf{0.090} & \textbf{24.28} & \textbf{0.831} & \textbf{0.116} & \textbf{29.88} & \textbf{0.861} & \textbf{0.097} \\
\hline
\end{tabular}
}
\renewcommand{\arraystretch}{1.0}
\vspace{-3pt}
\end{table*}

\begin{table}[t]
\centering
\caption{Quantitative comparison for video frame prediction and interpolation on the synthetic event dataset: GoPro~\cite{nah2017deep}. The best and second-best results are highlighted in \textbf{bold} and \underline{underline}, respectively.}
\label{tab:main_results_gopro}
\small
\renewcommand{\arraystretch}{1.15}
\resizebox{0.85\columnwidth}{!}{
\begin{tabular}{c|c|ccc|ccc}
\hline
\multirow{3}{*}{Task} & \multirow{3}{*}{Method} &
\multicolumn{6}{c}{GoPro~\cite{nah2017deep}} \\
\cline{3-8}
& & \multicolumn{3}{c|}{7 frames} & \multicolumn{3}{c}{15 frames} \\
\cline{3-8}
& & PSNR$\uparrow$ & SSIM$\uparrow$ & LPIPS$\downarrow$ &
PSNR$\uparrow$ & SSIM$\uparrow$ & LPIPS$\downarrow$ \\
\hline
\hline
\multirow{2}{*}{\makecell[c]{Frame\\Prediction}}
& VFPSIE~\cite{zhu2024video} (AAAI'24)
& \underline{18.26} & \underline{0.557} & \underline{0.386} & \underline{17.53} & \underline{0.523} & \underline{0.438} \\
& \textbf{\ours (VFP)}
& \textbf{20.87} & \textbf{0.703} & \textbf{0.231} & \textbf{19.75} & \textbf{0.672} & \textbf{0.308} \\
\hline
\multirow{5}{*}{\makecell[c]{Frame\\Interpolation}}
& TimeLens~\cite{tulyakov2021time} (CVPR'21)
& 17.29 & 0.571 & 0.300 & 15.02 & 0.531 & 0.384 \\
& CBMNet-Large~\cite{kim2023cbmnet} (CVPR'23)
& 17.98 & 0.592 & 0.327 & 17.44 & 0.532 & 0.372 \\
& RE-VDM~\cite{chen2025repurposing} (CVPR'25)
& \textbf{21.28} & \textbf{0.721} & \textbf{0.200} & \textbf{20.21} & \textbf{0.683} & \textbf{0.214} \\
& EGVD~\cite{zhang2025egvd} (ICCVW'25)
& 15.82 & 0.586 & 0.454 & 14.01 & 0.514 & 0.484 \\
& \textbf{\ours (VFI)}
& \underline{20.93} & \underline{0.711} & \underline{0.217} & \underline{19.77} & \underline{0.662} & \underline{0.301} \\
\hline
\end{tabular}
}
\renewcommand{\arraystretch}{1.0}
\vspace{-3pt}
\end{table}

\subsection{Implementation Details}
\label{subsec:implementation}
We build our framework upon Stable Diffusion~2.1~\cite{rombach2022high}, freezing the weights of both the VAE and U-net.
In stage 1, we train ER-VAE. 
In stage 2, we fine-tune the ControlNet, initialized by copying the parameters of the original U-net, and a $4\times4$ grid-based Latent Tokenizer that encodes spatial latent tokens. 
As Stable Diffusion~2.1 does not include an image encoder, we use the CLIP image encoder pretrained on the LAION-2B dataset~\cite{schuhmann2022laion} for conditional input construction.
The anchor image is encoded into a $1024$-dimensional embedding. 
The embedding is concatenated with the anchor latent tokens generated by the Latent Tokenizer and provided as conditioning through the cross-attention layers of the diffusion model.
For input preprocessing, images are divided into patches, and each patch is independently processed during generation. 
The event frames undergo the same spatial alignment process. 
The final frame is reconstructed by merging overlapping regions through a weighted average of patch predictions, enabling large-scale frame reconstruction.
Optimization is performed using AdamW~\cite{loshchilov2017decoupled} with a learning rate of $5\times10^{-4}$, a weight decay coefficient of $1\times10^{-2}$, and default hyperparameter settings for momentum and stability terms.

We evaluate reconstruction quality using PSNR~\cite{gonzalez2009digital}, SSIM~\cite{wang2004image}, and LPIPS~\cite{zhang2018unreasonable}.
We compare \ours\ with representative task-specific baselines: VFPSIE~\cite{zhu2024video} for video frame prediction, and TimeLens~\cite{tulyakov2021time}, CBMNet-Large~\cite{kim2023cbmnet}, RE-VDM~\cite{chen2025repurposing}, and EGVD~\cite{zhang2025egvd} for video frame interpolation.
Additional comparisons against other baselines using different input modalities and task formulations, including RGB-only frame prediction, event-only frame reconstruction, and event-based optical-flow-and-warping methods, are provided in the Supplementary Material.

\begin{figure*}[t!]
\centering
\includegraphics[width=1.0\textwidth]{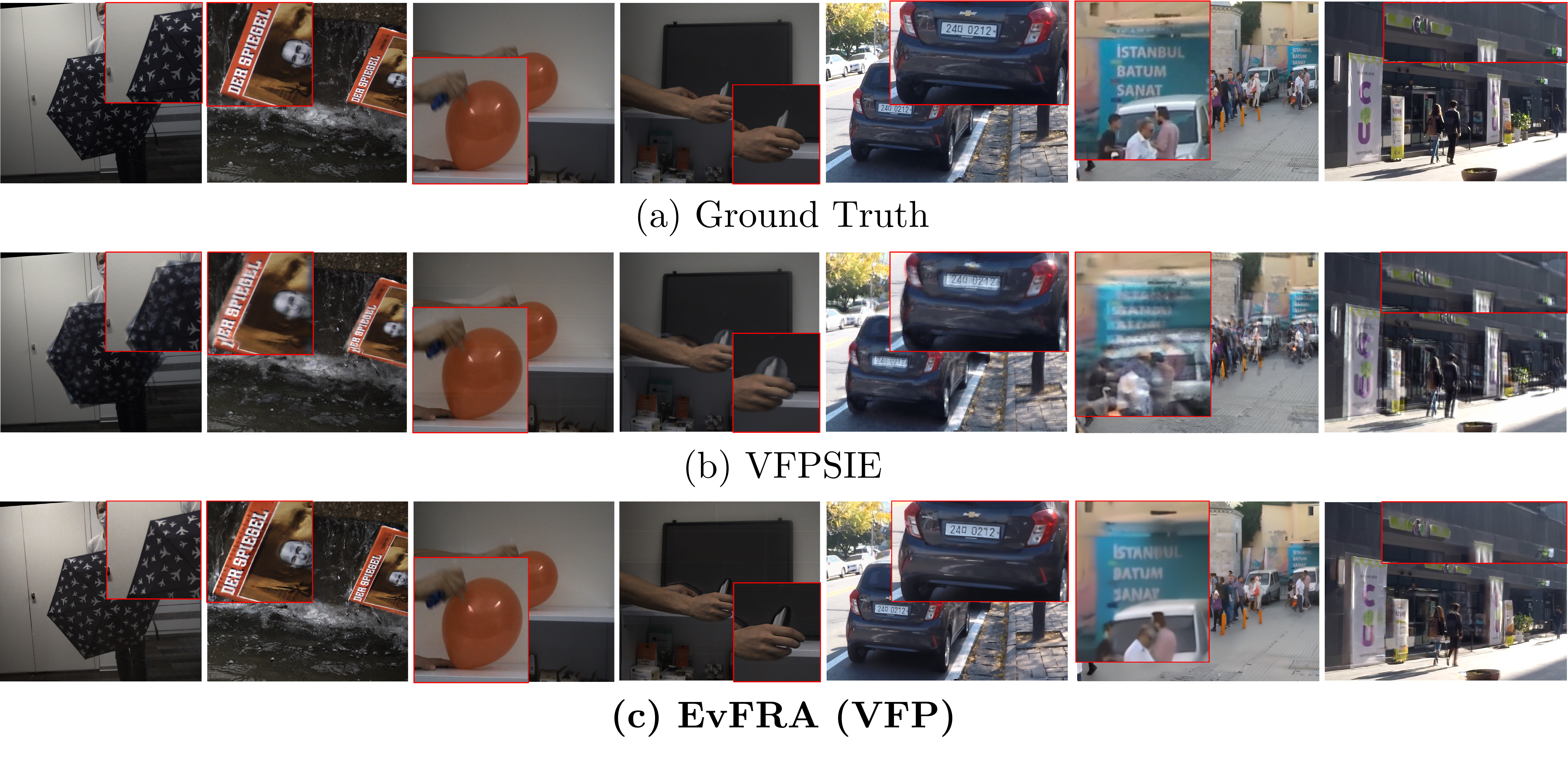}
\caption{Qualitative comparison for video frame prediction on HS-ERGB~\cite{tulyakov2021time} and GoPro~\cite{nah2017deep}.
Compared with VFPSIE~\cite{zhu2024video}, \ours\ better preserves moving object boundaries and text details while reducing blur and scattered motion artifacts.}
\label{fig:vfp_results}
\end{figure*}

\begin{figure*}[t!]
\includegraphics[width=1.0\textwidth]{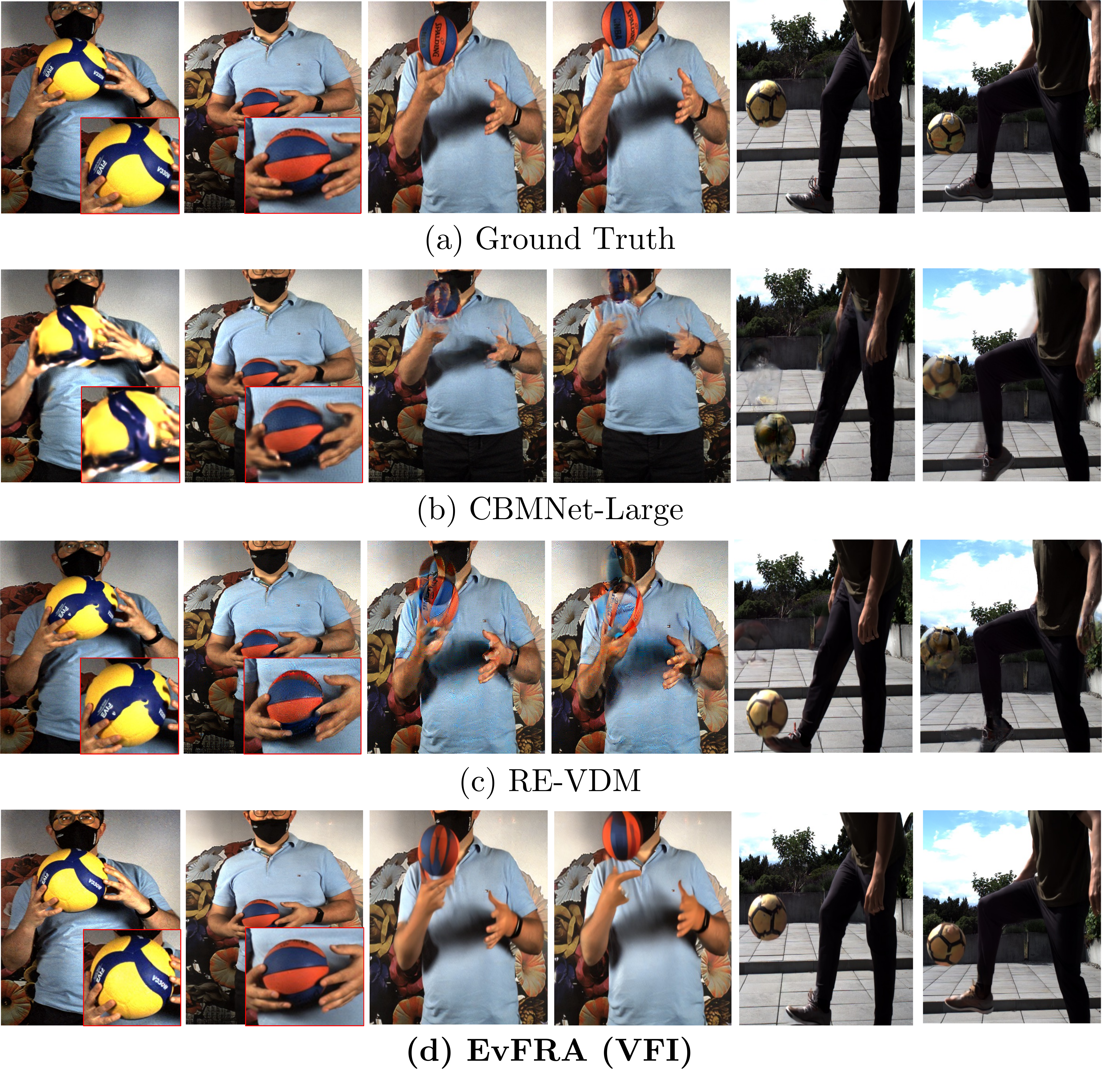}
\caption{Qualitative comparison for video frame interpolation on BS-ERGB~\cite{tulyakov2022timelenspp}.
Compared with CBMNet-Large~\cite{kim2023cbmnet} and RE-VDM~\cite{chen2025repurposing}, \ours\ more accurately reconstructs rapidly moving objects with fewer duplicated contours, geometric distortions, and motion blur.}
\label{fig:vfi_results}
\end{figure*}

\subsection{Results}
\label{subsec:results}
\textbf{Quantitative Results.}
As shown in Tables~\ref{tab:main_results_real} and~\ref{tab:main_results_gopro}, \ours\ achieves strong performance for both video frame prediction and interpolation.
On BS-ERGB~\cite{tulyakov2022timelenspp} and HS-ERGB~\cite{tulyakov2021time}, our frame prediction method achieves the best PSNR and SSIM across all intervals.
For video frame interpolation, our method achieves the best results across all metrics on both real-world datasets, while ranking second-best across all metrics on GoPro~\cite{nah2017deep}.
Overall, the results demonstrate that aligning event representations with anchor-to-target residual latents enables effective frame synthesis across prediction and interpolation settings.

\noindent\textbf{Qualitative Results.}
Figure~\ref{fig:vfp_results} shows that, compared with VFPSIE~\cite{zhu2024video}, our frame prediction method better preserves object boundaries, fine structures, and text while reducing blur and motion artifacts on HS-ERGB~\cite{tulyakov2021time} and GoPro~\cite{nah2017deep}.
For frame interpolation, Figure~\ref{fig:vfi_results} shows that ours produces fewer duplicated contours, geometric distortions, and motion blur around rapidly moving objects than CBMNet-Large~\cite{kim2023cbmnet} and RE-VDM~\cite{chen2025repurposing} on BS-ERGB~\cite{tulyakov2022timelenspp}.

\noindent\textbf{Video Frame Consistency.}
We evaluate how consistent the frame synthesized by our method is with the existing frames.
Although VBench~\cite{huang2023vbench} is widely used for temporal consistency evaluation in video generation, it can be unreliable in our scene synthesis setting because it may assign high scores to overly smooth or static outputs.
Therefore, we compute optical flow between consecutive synthesized and ground-truth frames using RAFT~\cite{teed2020raft} and report the endpoint error between the resulting motion fields in Table~\ref{tab:temporal_consistency}.
\ours\ achieves the lowest endpoint error across all datasets for both frame prediction and interpolation, indicating more accurate motion reproduction and better temporal consistency.

\subsection{Ablation Studies}
\label{subsec:ablation}
\textbf{VLT Augmentation.}
We compare VLT Augmentation against fixed single-interval training.
By progressively increasing the prediction interval from 1 to 2 and then 3 frames, VLT Augmentation consistently improves performance across both datasets, as shown in Table~\ref{tab:ablation_vlt}.
Additional ablations on the curriculum order and interval-sampling ratios are provided in the Supplementary Material. \\

\begin{table*}[t]
\centering
\begin{minipage}{0.48\textwidth}
\centering
\caption{Evaluation of temporal consistency for video frame prediction (VFP) and interpolation (VFI) on BS-ERGB~\cite{tulyakov2022timelenspp} (3-frame), HS-ERGB~\cite{tulyakov2021time} (7-frame), and GoPro~\cite{nah2017deep} (15-frame).}
\label{tab:temporal_consistency}
\footnotesize
\renewcommand{\arraystretch}{1.1}
\resizebox{\columnwidth}{!}{
\begin{tabular}{c|ccc}
\hline
\multirow{2}{*}{Method}
& BS-ERGB~\cite{tulyakov2022timelenspp}
& HS-ERGB~\cite{tulyakov2021time}
& GoPro~\cite{nah2017deep} \\
\cline{2-4}
& \multicolumn{3}{c}{Optical Flow Endpoint Error $\downarrow$} \\
\hline
VFPSIE~\cite{zhu2024video}
& 5.509 & 3.685 & 5.239 \\
\textbf{\ours (VFP)}
& \textbf{5.092} & \textbf{3.350} & \textbf{5.002} \\
\hline
TimeLens~\cite{tulyakov2021time}
& 5.311 & 3.713 & 5.309 \\
CBMNet-Large~\cite{kim2023cbmnet}
& 6.079 & 4.109 & 5.321 \\
RE-VDM~\cite{chen2025repurposing}
& 6.106 & 4.201 & 5.318 \\
\textbf{\ours (VFI)}
& \textbf{4.612} & \textbf{3.002} & \textbf{4.899} \\
\hline
\end{tabular}
}
\renewcommand{\arraystretch}{1.0}
\normalsize
\end{minipage}
\hfill
\begin{minipage}{0.48\textwidth}
\centering
\caption{Ablation on VLT Augmentation. We sequentially train the model with skip-frame lengths of 1, 2, and 3. Results are reported on HS-ERGB~\cite{tulyakov2021time} (7-frame) and GoPro~\cite{nah2017deep} (15-frame) prediction settings.}
\label{tab:ablation_vlt}
\footnotesize
\resizebox{\columnwidth}{!}{
\begin{tabular}{c|ccc|ccc}
\hline
& \multicolumn{3}{c|}{HS-ERGB~\cite{tulyakov2021time}} &
\multicolumn{3}{c}{GoPro~\cite{nah2017deep}} \\
\cline{2-7}
& PSNR$\uparrow$ & SSIM$\uparrow$ & LPIPS$\downarrow$ &
PSNR$\uparrow$ & SSIM$\uparrow$ & LPIPS$\downarrow$ \\
\hline
w/o & 24.87 & 0.816 & 0.255 & 14.09 & 0.556 & 0.533 \\
with & \textbf{29.74} & \textbf{0.858} & \textbf{0.101} & \textbf{19.75} & \textbf{0.672} & \textbf{0.308} \\
\hline
\end{tabular}
}
\normalsize
\end{minipage}
\end{table*}

\begin{table*}[t]
\centering
\begin{minipage}{0.48\textwidth}
\centering
\caption{Analysis of stage contributions. 
$\mathcal{D}$ denotes the pretrained VAE decoder. 
Direct decoding with $z_{\mathrm{event}}^{\mathrm{VAE}}$ evaluates the unaligned event latent, direct decoding with $z_{\mathrm{event}}^{\mathrm{ER}}$ evaluates the stage~1 aligned latent, and \ours\ denotes the full pipeline with stage~2 residual denoising.}
\label{tab:stage1_vs_stage2}
\small
\renewcommand{\arraystretch}{1.15}
\resizebox{\columnwidth}{!}{
\begin{tabular}{c|ccc}
\hline
Metric
& $\mathcal{D}(z_t + z^{\mathrm{VAE}}_{\mathrm{event}})$
& $\mathcal{D}(z_t + z^{\mathrm{ER}}_{\mathrm{event}})$
& \ours \\
\hline
PSNR$\uparrow$ & 20.23 & 23.80 & \textbf{24.20} \\
SSIM$\uparrow$ & 0.652 & 0.706 & \textbf{0.804} \\
\hline
\end{tabular}
}
\renewcommand{\arraystretch}{1.0}
\normalsize
\end{minipage}
\hfill
\begin{minipage}{0.48\textwidth}
\centering
\caption{Ablation of event conditioning strategies on BS-ERGB~\cite{tulyakov2022timelenspp} under the 3-frame video frame prediction setting.}
\label{tab:conditioning_strategy}
\small
\renewcommand{\arraystretch}{1.15}
\resizebox{0.78\columnwidth}{!}{
\begin{tabular}{c|ccc}
\hline
Conditioning & PSNR$\uparrow$ & SSIM$\uparrow$ & LPIPS$\downarrow$ \\
\hline
ControlNet & \textbf{24.20} & \textbf{0.804} & \textbf{0.119} \\
Latent concat. & 19.62 & 0.603 & 0.244 \\
Cross-attn only & 21.97 & 0.707 & 0.178 \\
\hline
\end{tabular}
}
\renewcommand{\arraystretch}{1.0}
\normalsize
\end{minipage}

\end{table*}

\begin{figure}[t]
    \centering
    \begin{subfigure}{0.49\columnwidth}
        \centering
        \includegraphics[width=\linewidth]{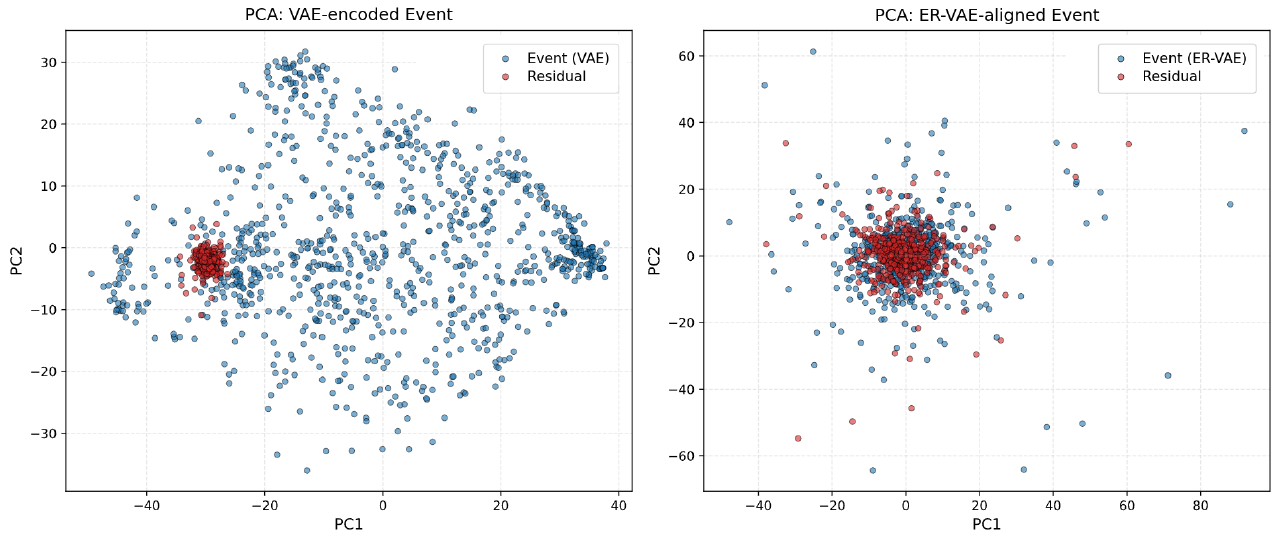}
        \caption{PCA visualization of VAE-encoded and ER-VAE-aligned event latents.}
        \label{fig:er_vae_pca}
    \end{subfigure}
    \hfill
    \begin{subfigure}{0.49\columnwidth}
        \centering
        \includegraphics[width=\linewidth]{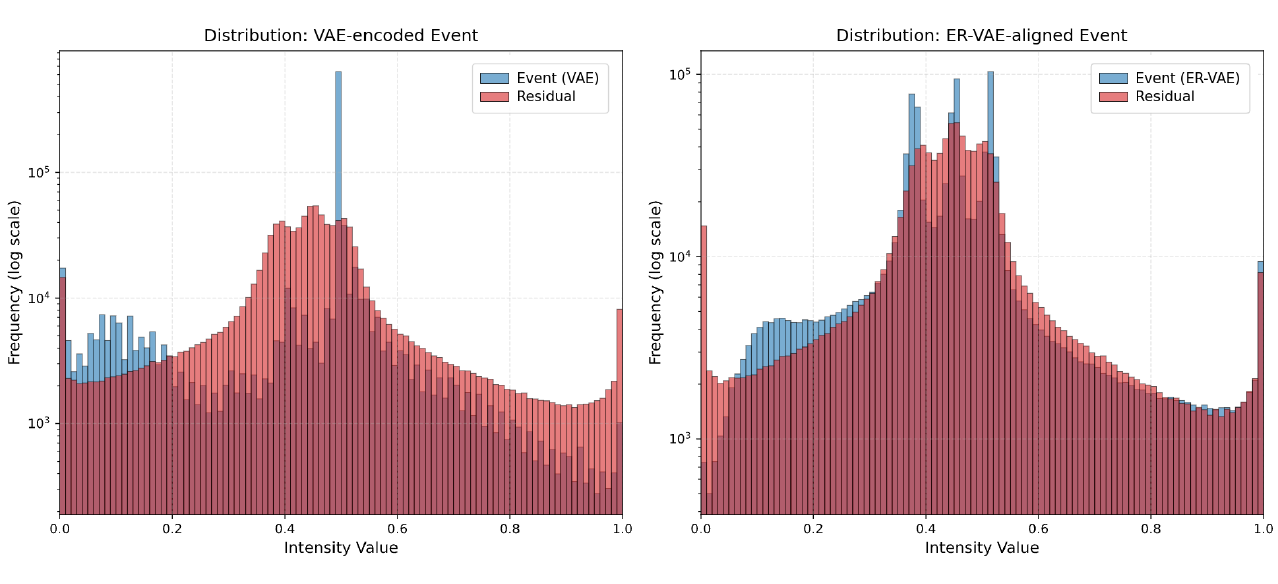}
        \caption{Intensity distributions of decoded event and residual latents.}
        \label{fig:er_vae_distribution}
    \end{subfigure}

    \caption{Analysis of the effectiveness of the ER-VAE in both the latent and pixel spaces.}
    \label{fig:er_vae_analysis}
\end{figure}

\begingroup
\linespread{1.06}\selectfont
\noindent\textbf{Stage Contributions.}
We analyze the contribution of each training stage on BS-ERGB~\cite{tulyakov2022timelenspp} under the 3-frame video frame prediction setting in Table~\ref{tab:stage1_vs_stage2}.
Here, $\mathcal{D}$ denotes the pretrained VAE decoder, and $z_{\mathrm{event}}^{\mathrm{VAE}}$ denotes the event latent encoded by the pretrained VAE.
$\mathcal{D}(z_t + z_{\mathrm{event}}^{\mathrm{VAE}})$ evaluates reconstruction without ER-VAE alignment, while $\mathcal{D}(z_t + z_{\mathrm{event}}^{\mathrm{ER}})$ evaluates the stage~1 aligned event latent without diffusion refinement.
The full \ours\ pipeline includes both stage~1 alignment and stage~2 residual denoising.
It appears that the two stages contribute in a complementary manner: Stage~1 provides large pixel-level alignment (PSNR improvement from 20.23 to 23.80 dB) by aligning events with the residual, while stage~2 yields large enhancement in the synthesized image structure (SSIM improvement from 0.706 to 0.804) through diffusion-based residual refinement. \\
\endgroup

\noindent\textbf{ER-VAE Analysis.}
To analyze the effect of ER-VAE, we examine its impact in both the latent and pixel spaces.
Figure~\ref{fig:er_vae_pca} shows that ER-VAE alignment transforms widely scattered event latents compared to the residual latents into a more compact distribution that better overlaps with the residual latent cluster.
Figure~\ref{fig:er_vae_distribution} further shows that the decoded ER-VAE-aligned event representation more closely matches the decoded residual latent across the intensity distribution.

\begin{figure}[t]
\centering
\setlength{\tabcolsep}{1.5pt}
\renewcommand{\arraystretch}{0.5}
\begin{tabular}{ccc}
\includegraphics[width=0.23\columnwidth]{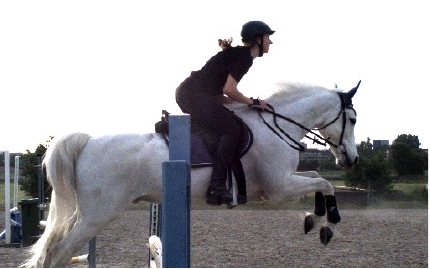} &
\includegraphics[width=0.23\columnwidth]{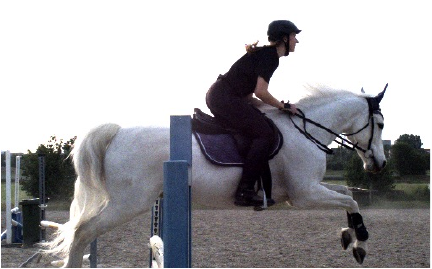} &
\includegraphics[width=0.23\columnwidth]{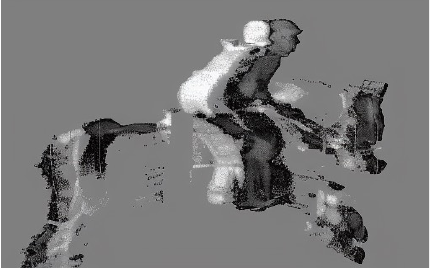} \\
{\scriptsize (a) Anchor} &
{\scriptsize (b) Target} &
{\scriptsize (c) Event Frame} \\[1pt]
\includegraphics[width=0.23\columnwidth]{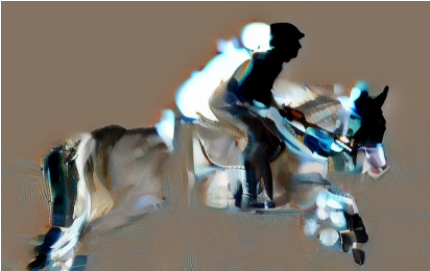} &
\includegraphics[width=0.23\columnwidth]{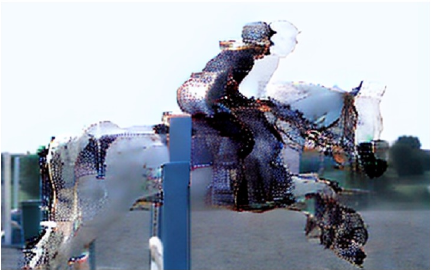} &
\includegraphics[width=0.23\columnwidth]{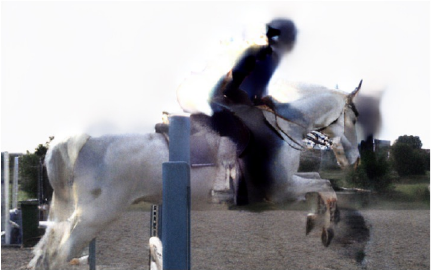} \\
{\scriptsize (d) $\mathcal{D}(z_{\mathrm{res}})$} &
{\scriptsize (e) $\mathcal{D}(z_t + z_{\mathrm{event}}^{\mathrm{VAE}})$} &
{\scriptsize (f) $\mathcal{D}(z_t + z_{\mathrm{event}}^{\mathrm{ER}})$}
\end{tabular}
\caption{Qualitative direct-decoding analysis of ER-VAE alignment.
$\mathcal{D}$ denotes the pretrained VAE decoder, and $z_{\mathrm{event}}^{\mathrm{VAE}}$ and $z_{\mathrm{event}}^{\mathrm{ER}}$ denote the VAE-encoded and ER-VAE-aligned event latents, respectively.
The second row compares (d) the decoded ground-truth residual latent with reconstructions from (e) the unaligned and (f) aligned event latents, showing that ER-VAE alignment reduces structural artifacts and produces a result closer to the target frame.}

\label{fig:er_vae_decoded}
\vspace{-14pt}
\end{figure}

In addition, Figure~\ref{fig:er_vae_decoded} provides qualitative analysis of ER-VAE alignment.
The event frame in (c) and the decoded residual latent in (d) exhibit corresponding structures around motion regions.
Compared with the severe structural artifacts produced by directly decoding the unaligned event latent in (e), the ER-VAE-aligned reconstruction in (f) better preserves the moving subject and scene structure. 
These results demonstrate that ER-VAE improves consistency in both the latent and decoded representation spaces, supporting the effectiveness of event-to-residual alignment. \\

\noindent\textbf{Event Conditioning Strategy.}
We compare ControlNet-based conditioning with latent concatenation and cross-attention-only conditioning under identical settings.
While latent concatenation injects events only at the input latent level and cross-attention-only conditioning lacks explicit spatial modulation, ControlNet injects event-conditioned spatial motion cues through trainable control branches.
As shown in Table~\ref{tab:conditioning_strategy}, ControlNet achieves the best performance across all metrics, highlighting the importance of structured event conditioning.

\section{Conclusion}
\label{sec:conclusion}
\vspace{-0.7em}
We introduced \ours, an event-based scene synthesis framework for video frame prediction and interpolation based on inter-frame residual alignment.
ER-VAE exploits the structural correspondence between events and frame-to-frame changes to align event representations with residual latents, providing an event-informed initialization for residual denoising.
VLT Augmentation further improves generalization across varying motion magnitudes and temporal intervals.
Extensive experiments demonstrated that \ours\ achieves consistently strong quantitative and qualitative performance in both video frame prediction and interpolation. \\
\noindent \textbf{Limitations.} Our current framework synthesizes one target frame per inference step, while extending the anchor-to-target formulation to jointly generate multiple target frames remains future work.
\vspace{-0.8em}
\section*{Acknowledgements}
\vspace{-0.5em}
\label{sec:acknowledgements}
Jiyun Kong and Jong-Seok Lee were supported by the National Research Foundation of Korea (NRF) grant (No. RS-2024-00453301, RS-2025-00517159) and by Institute of Information \& communications Technology Planning \& Evaluation (IITP) grant (IITP-2025-RS-2024-00428780).
Jun-Hyuk Kim was supported by the Institute of Information \& Communications Technology Planning \& Evaluation (IITP) grant funded by the Korea government (MSIT) [RS-2021-II211341, Artificial Intelligence Graduate School Program (Chung-Ang University)].

\bibliographystyle{splncs04}
\bibliography{main}

@String(AAAI  = {AAAI})

@article{lichtsteiner2008dvs,
  title={A 128$\times$128 120 db 15$\mu$s latency asynchronous temporal contrast vision sensor},
  author={Lichtsteiner, P. and Posch, C. and Delbruck, T.},
  journal={IEEE Journal of Solid-State Circuits},
  year={2008},
  publisher={IEEE}
}

@inproceedings{chakravarthi2024recent,
  title={Recent event camera innovations: A survey},
  author={Chakravarthi, B. and Verma, A. A. and Daniilidis, K. and Fermuller, C. and Yang, Y.},
  booktitle={European Conference on Computer Vision},
  year={2024}
}

@inproceedings{zhu2024video,
  title={Video frame prediction from a single image and events},
  author={Zhu, J. and Wan, Z. and Dai, Y.},
  booktitle={Proceedings of the AAAI Conference on Artificial Intelligence},
  year={2024}
}

@inproceedings{wang2025event,
  title={Event-based continuous color video decompression from single frames},
  author={Wang, Z. and Hamann, F. and Chaney, K. and Jiang, W. and Gallego, G. and Daniilidis, K.},
  booktitle={Proceedings of the IEEE/CVF International Conference on Computer Vision},
  year={2025}
}

@article{gehrig2024dense,
  title={Dense continuous-time optical flow from event cameras},
  author={Gehrig, M. and Muglikar, M. and Scaramuzza, D.},
  journal={IEEE Transactions on Pattern Analysis and Machine Intelligence},
  year={2024},
  publisher={IEEE}
}

@inproceedings{gehrig2021eraft,
  title={E-raft: Dense optical flow from event cameras},
  author={Gehrig, M. and Millhäusler, M. and Gehrig, D. and Scaramuzza, D.},
  booktitle={2021 International Conference on 3D Vision (3DV)},
  year={2021}
}

@article{shiba2024secrets,
  title={Secrets of event-based optical flow, depth and ego-motion estimation by contrast maximization},
  author={Shiba, S. and Klose, Y. and Aoki, Y. and Gallego, G.},
  journal={IEEE Transactions on Pattern Analysis and Machine Intelligence},
  year={2024},
  publisher={IEEE}
}

@inproceedings{ni2023conditional,
  title={Conditional image-to-video generation with latent flow diffusion models},
  author={Ni, H. and Shi, C. and Li, K. and Huang, S. X. and Min, M. R.},
  booktitle={Proceedings of the IEEE/CVF Conference on Computer Vision and Pattern Recognition},
  year={2023}
}

@inproceedings{blattmann2023videoldm,
  title={Align your latents: High-resolution video synthesis with latent diffusion models},
  author={Blattmann, A. and Rombach, R. and Ling, H. and Dockhorn, T. and Kim, S. W. and Fidler, S. and Kreis, K.},
  booktitle={Proceedings of the IEEE/CVF Conference on Computer Vision and Pattern Recognition},
  year={2023}
}

@inproceedings{rombach2022high,
  title={High-resolution image synthesis with latent diffusion models},
  author={Rombach, R. and Blattmann, A. and Lorenz, D. and Esser, P. and Ommer, B.},
  booktitle={Proceedings of the IEEE/CVF Conference on Computer Vision and Pattern Recognition},
  year={2022}
}

@inproceedings{peebles2023scalable,
  title={Scalable diffusion models with transformers},
  author={Peebles, W. and Xie, S.},
  booktitle={Proceedings of the IEEE/CVF International Conference on Computer Vision},
  year={2023}
}

@inproceedings{ho2020denoising,
  title={Denoising diffusion probabilistic models},
  author={Ho, J. and Jain, A. and Abbeel, P.},
  booktitle={Advances in Neural Information Processing Systems},
  year={2020}
}

@article{ho2022cascaded,
  title={Cascaded diffusion models for high fidelity image generation},
  author={Ho, J. and Saharia, C. and Chan, W. and Fleet, D. J. and Norouzi, M. and Salimans, T.},
  journal={Journal of Machine Learning Research},
  year={2022}
}

@inproceedings{liu2017voxel,
  title={Video frame synthesis using deep voxel flow},
  author={Liu, Z. and Yeh, R. A. and Tang, X. and Liu, Y. and Agarwala, A.},
  booktitle={Proceedings of the IEEE International Conference on Computer Vision},
  year={2017}
}

@inproceedings{liang2024flowvid,
  title={Flowvid: Taming imperfect optical flows for consistent video-to-video synthesis},
  author={Liang, F. and Wu, B. and Wang, J. and Yu, L. and Li, K. and Zhao, Y.},
  booktitle={Proceedings of the IEEE/CVF Conference on Computer Vision and Pattern Recognition},
  year={2024}
}

@inproceedings{zhang2024extdm,
  title={Extdm: Distribution extrapolation diffusion model for video prediction},
  author={Zhang, Z. and Hu, J. and Cheng, W. and Paudel, D. and Yang, J.},
  booktitle={Proceedings of the IEEE/CVF Conference on Computer Vision and Pattern Recognition},
  year={2024}
}

@inproceedings{bardow2016simultaneous,
  title={Simultaneous optical flow and intensity estimation from an event camera},
  author={Bardow, P. and Davison, A. J. and Leutenegger, S.},
  booktitle={Proceedings of the IEEE Conference on Computer Vision and Pattern Recognition},
  year={2016}
}

@inproceedings{rebecq2019events,
  title={Events-to-video: Bringing modern computer vision to event cameras},
  author={Rebecq, H. and Ranftl, R. and Koltun, V. and Scaramuzza, D.},
  booktitle={Proceedings of the IEEE/CVF Conference on Computer Vision and Pattern Recognition},
  year={2019}
}

@article{ercan2024hypere2vid,
  title={Hypere2vid: Improving event-based video reconstruction via hypernetworks},
  author={B. Ercan and O. Eker and C. Saglam and A. Erdem and E. Erdem},
  journal={IEEE Transactions on Image Processing},
  year={2024},
  publisher={IEEE}
}

@inproceedings{qu2024e2hqv,
  title={E2hqv: High-quality video generation from event camera via theory-inspired model-aided deep learning},
  author={Qu, Q. and Shen, Y. and Chen, X. and Chung, Y. Y. and Liu, T.},
  booktitle={Proceedings of the AAAI Conference on Artificial Intelligence},
  year={2024}
}

@inproceedings{zhou2021curriculum,
  title={Curriculum learning by optimizing learning dynamics},
  author={Zhou, T. and Wang, S. and Bilmes, J.},
  booktitle={Proceedings of the International Conference on Artificial Intelligence and Statistics},
  year={2021},

}

@inproceedings{hu2023dmvfn,
  title={A dynamic multi-scale voxel flow network for video prediction},
  author={Hu, X. and Huang, Z. and Huang, A. and Xu, J. and Zhou, S.},
  booktitle={Proceedings of the IEEE/CVF Conference on Computer Vision and Pattern Recognition},
  year={2023}
}

@inproceedings{tulyakov2022timelenspp,
  title={Time lens++: Event-based frame interpolation with parametric non-linear flow and multi-scale fusion},
  author={Tulyakov, S. and Bochicchio, A. and Gehrig, D. and Georgoulis, S. and Li, Y. and Scaramuzza, D.},
  booktitle={Proceedings of the IEEE/CVF Conference on Computer Vision and Pattern Recognition},
  year={2022}
}

@article{gao2022superfast,
  title={Superfast: 200× Video frame interpolation via event camera},
  author={Kim, T. and Chae, Y. and Jang, H. K. and Yoon, K. J.},
  journal={IEEE Transactions on Pattern Analysis and Machine Intelligence},
  year={2022}
}

@inproceedings{kim2023cbmnet,
  title={Event-based video frame interpolation with cross-modal asymmetric bidirectional motion fields},
  author={Kim, T. and Chae, Y. and Jang, H. K. and Yoon, K. J. },
  booktitle={Proceedings of the IEEE/CVF Conference on Computer Vision and Pattern Recognition},
  year={2023}
}

@inproceedings{jeong2024ocai,
  title={Ocai: Improving optical flow estimation by occlusion and consistency aware interpolation},
  author={Jeong, J. and Cai, H. and Garrepalli, R. and Lin, J. M. and Hayat, M. and Porikli, F.},
  booktitle={Proceedings of the IEEE/CVF Conference on Computer Vision and Pattern Recognition},
  year={2024}
}

@inproceedings{barhaim2020scopeflow,
  title={Scopeflow: Dynamic scene scoping for optical flow},
  author={Bar-Haim, A. and Wolf, L.},
  booktitle={Proceedings of the IEEE/CVF Conference on Computer Vision and Pattern Recognition},
  year={2020}
}

@article{wan2022learning,
  title={Learning dense and continuous optical flow from an event camera},
  author={Wan, Z. and Dai, Y. and Mao, Y.},
  journal={IEEE Transactions on Image Processing},
  year={2022},
  publisher={IEEE}
}

@inproceedings{sohl2015deep,
  title={Deep unsupervised learning using nonequilibrium thermodynamics},
  author={Sohl-Dickstein, J. and Weiss, E. and Maheswaranathan, N. and Ganguli, S.},
  booktitle={International Conference on Machine Learning},
  year={2015}
}

@inproceedings{dhariwal2021diffusion,
  title={Diffusion models beat gans on image synthesis},
  author={Dhariwal, P. and Nichol, A.},
  booktitle={Advances in Neural Information Processing Systems},
  year={2021}
}

@inproceedings{zhang2023adding,
  title={Adding conditional control to text-to-image diffusion models},
  author={Zhang, L. and Agrawala, M.},
  booktitle={Proceedings of the IEEE/CVF International Conference on Computer Vision},
  year={2023}
}

@inproceedings{saxena2024surprising,
  title={The surprising effectiveness of diffusion models for optical flow and monocular depth estimation},
  author={Saxena, S. and Herrmann, C. and Hur, J. and Kar, A. and Norouzi, M. and Sun, D. and Fleet, D. J.},
  booktitle={Advances in Neural Information Processing Systems},
  volume={36},
  year={2024}
}

@misc{chen2023controlavideo,
  title={Control-A-Video: Controllable Text-to-Video Generation with Diffusion Models}, 
  author={Chen, W. and Ji, Y. and Wu, J. and Wu, H. and Xie, P. and Li, J. and Xia, X. and Xiao, X. and Lin, L.},
  year={2023},
  eprint={2305.13840},
  archivePrefix={arXiv},
  primaryClass={cs.CV}
}

@inproceedings{wu2024emotion,
  title={E-motion: Future motion simulation via event sequence diffusion},
  author={Wu, S. and Zhu, Z. and Hou, J. and Shi, G. and Wu, J.},
  booktitle={Advances in Neural Information Processing Systems},
  year={2024}
}

@inproceedings{zhang2025egvd,
  title={Egvd: Event-guided video diffusion model for physically realistic large-motion frame interpolation},
  author={Zhang, Z. and Li, X. and Liu, Y. and Wang, Y. and Chen, Y. and Xue, T. and Guo, S.},
  booktitle={Proceedings of the IEEE/CVF International Conference on Computer Vision Workshops},
  year={2025}
}

@inproceedings{chen2025repurposing,
  title={Repurposing pre-trained video diffusion models for event-based video interpolation},
  author={Chen, J. and Feng, B. Y. and Cai, H. and Wang, T. and Burner, L. and Yuan, D. and Fermuller, C. and Metzler, C. A. and Aloimonos, Y.},
  booktitle={Proceedings of the IEEE/CVF Conference on Computer Vision and Pattern Recognition},
  year={2025}
}

@inproceedings{kingma2013auto,
  title={Auto-encoding variational bayes},
  author={Kingma, D. P. and Welling, M.},
  booktitle={International Conference on Learning Representations},
  year={2014}
}

@article{ghosh2019variational,
  title={From variational to deterministic autoencoders},
  author={Ghosh, P. and Sajjadi, M. S. M. and Vergari, A. and Black, M. and Sch{\"o}lkopf, B.},
  journal={arXiv preprint arXiv:1903.12436},
  year={2019}
}

@inproceedings{karras2022elucidating,
  title={Elucidating the design space of diffusion-based generative models},
  author={Karras, T. and Aittala, M. and Aila, T. and Laine, S.},
  booktitle={Advances in neural information processing systems},
  year={2022}
}

@inproceedings{zhang2018unreasonable,
  title={The unreasonable effectiveness of deep features as a perceptual metric},
  author={Zhang, R. and Isola, P. and Efros, A. A. and Shechtman, E. and Wang, O.},
  booktitle={Proceedings of the IEEE Conference on Computer Vision and Pattern Recognition},
  year={2018}
}

@article{gallego2020event,
  title={Event-based vision: A survey},
  author={Gallego, G. and Delbr{\"u}ck, T. and Orchard, G. and Bartolozzi, C. and Taba, B. and Censi, A. and Leutenegger, S. and Davison, A. J. and Conradt, J. and Daniilidis, K. and Scaramuzza, D.},
  journal={IEEE transactions on pattern analysis and machine intelligence},
  year={2020},
  publisher={IEEE}
}

@inproceedings{radford2021learning,
  title={Learning transferable visual models from natural language supervision},
  author={Radford, A. and Kim, J. W. and Hallacy, C. and Ramesh, A. and Goh, G. and Agarwal, S. and Sastry, G. and Askell, A. and Mishkin, P. and Clark, J. and Krueger, G. and Sutskever, I.},
  booktitle={International Conference on Machine Learning},
  year={2021}
}

@inproceedings{tulyakov2021time,
  title={Time lens: Event-based video frame interpolation},
  author={Tulyakov, S. and Gehrig, D. and Georgoulis, S. and Erbach, J. and Gehrig, M. and Li, Y. and Scaramuzza, D.},
  booktitle={Proceedings of the {IEEE}/CVF Conference on Computer Vision and Pattern Recognition},
  year={2021}
}

@inproceedings{nah2017deep,
  title={Deep multi-scale convolutional neural network for dynamic scene deblurring},
  author={Nah, S. and Kim, T. H. and Lee, K. M.},
  booktitle={Proceedings of the IEEE/CVF Conference on Computer Vision and Pattern Recognition},
  year={2017}
}

@inproceedings{rebecq2018esim,
  title={Esim: An open event camera simulator},
  author={Rebecq, H. and Gehrig, D. and Scaramuzza, D.},
  booktitle={Conference on Robot Learning},
  year={2018}
}

@inproceedings{schuhmann2022laion,
  title={Laion-5b: An open large-scale dataset for training next generation image-text models},
  author={Schuhmann, C. and Beaumont, R. and Vencu, R. and Gordon, C. and Wightman, R. and Cherti, M. and Coombes, T. and Katta, A. and Mullis, C. and Wortsman, M. and Schramowski, P. and Kundurthy, S. and Crowson, K. and Schmidt, L. and Kaczmarczyk, R. and Jitsev, J.},
  booktitle={Advances in Neural Information Processing Systems},
  year={2022}
}

@article{loshchilov2017decoupled,
  title={Decoupled weight decay regularization},
  author={Loshchilov, I. and Hutter, F.},
  journal={arXiv preprint arXiv:1711.05101},
  year={2017}
}

@book{gonzalez2009digital,
  title={Digital image processing},
  author={Gonzalez, R. C. and Woods, R. E.},
  publisher={Pearson Education India},
  year={2009}
}

@article{wang2004image,
  title={Image quality assessment: From error visibility to structural similarity},
  author={Wang, Z. and Bovik, A. C. and Sheikh, H. R. and Simoncelli, E. P.},
  journal={IEEE Transactions on Image Processing},
  year={2004}
}

@InProceedings{huang2023vbench,
  title={Vbench: Comprehensive benchmark suite for video generative models},
  author={Huang, Z. and He, Y. and Yu, J. and Zhang, F. and Si, C. and Jiang, Y. and Zhang, Y. and Wu, T. and Jin, Q. and Chanpaisit, N. and Wang, Y. and Chen, X. and Wang, L. and Lin, D. and Qiao, Y. and Liu, Z.},
  booktitle={Proceedings of the IEEE/CVF Conference on Computer Vision and Pattern Recognition},
  year={2024}
}

@inproceedings{teed2020raft,
  title={Raft: Recurrent all-pairs field transforms for optical flow},
  author={Teed, Z. and Deng, J.},
  booktitle={European Conference on Computer Vision},
  year={2020}
}

@article{zheng2025eventdiff,
  title={EventDiff: A Unified and Efficient Diffusion Model Framework for Event-based Video Frame Interpolation},
  author={Zheng, Hanle and Han, Xujie and Peng, Zegang and Zhang, Shangbin and Du, Guangxun and Zou, Zhuo and Wang, Xilin and Wu, Jibin and Guo, Hao and Deng, Lei},
  journal={arXiv preprint arXiv:2505.08235},
  year={2025}
}

@inproceedings{jiang2018super,
  title={Super SloMo: High Quality Estimation of Multiple Intermediate Frames for Video Interpolation},
  author={Jiang, Huaizu and Sun, Deqing and Jampani, Varun and Yang, Ming-Hsuan and Learned-Miller, Erik and Kautz, Jan},
  booktitle={Proceedings of the IEEE/CVF Conference on Computer Vision and Pattern Recognition},
  year={2018}
}

@inproceedings{bao2019depth,
  title={Depth-Aware Video Frame Interpolation},
  author={Bao, Wenbo and Lai, Wei-Sheng and Ma, Chao and Zhang, Xiaoyun and Gao, Zhiyong and Yang, Ming-Hsuan},
  booktitle={Proceedings of the IEEE/CVF Conference on Computer Vision and Pattern Recognition},
  year={2019}
}

@inproceedings{reda2022film,
  title={FILM: Frame Interpolation for Large Motion},
  author={Reda, Fitsum and Kontkanen, Janne and Tabellion, Eric and Sun, Deqing and Pantofaru, Caroline and Curless, Brian},
  booktitle={European Conference on Computer Vision},
  year={2022}
}

@article{sun2026transient,
  title={Transient Video Interpolation for Dynamic Non-Line-of-Sight Imaging},
  author={Sun, S. and Li, Y. and Fu, J. and Xu, F. and Xiong, Z.},
  journal={Optics Express},
  year={2026}
}

@article{kai2026evtexturepp,
  title={EvTexture++: Event-Driven Texture Enhancement for Video Super-Resolution},
  author={Kai, D. and Lu, J. and Zhang, Y. and Sun, X.},
  journal={arXiv preprint arXiv:2606.13580},
  year={2026}
}

@article{ren2026systematic,
  title={A Systematic Survey on Event Camera Representation Learning},
  author={Ren, H. and Jiang, Y. and Huang, T. and Wu, X.},
  journal={arXiv preprint arXiv:2606.23078},
  year={2026}
}

@article{hoppe2022ramvid,
  title={Diffusion Models for Video Prediction and Infilling},
  author={H{\"o}ppe, Tobias and Mehrjou, Arash and Bauer, Stefan and Nielsen, Didrik and Dittadi, Andrea},
  journal={Transactions on Machine Learning Research},
  year={2022}
}

\clearpage
\appendix

\begin{center}
\Large{\textbf{Event-based Scene Synthesis via Inter-Frame Residual Alignment}} \\
\vspace{0.5em}
\large{Supplementary Material}
\end{center}

\renewcommand{\thesection}{\Alph{section}}
\renewcommand{\thesubsection}{\Alph{section}.\arabic{subsection}}
\renewcommand{\theHsection}{supp.\Alph{section}}
\renewcommand{\theHsubsection}{supp.\Alph{section}.\arabic{subsection}}
\renewcommand{\thetable}{S\arabic{table}}
\renewcommand{\thefigure}{S\arabic{figure}}

\setcounter{section}{0}
\setcounter{table}{0}
\setcounter{figure}{0}

\begin{center}
\large\textbf{Contents}
\end{center}
\vspace{0.42em}

\noindent A. \textbf{Additional Implementation Details} \dotfill \pageref{sec:implementation_details}\\[0.4em]
\noindent B. \textbf{Ablation Studies} \dotfill \pageref{sec:ablation_studies}\\[0.4em]
\makebox[1.5em][l]{}B.1. \textbf{Model Formulation and Event Conditioning} \dotfill 
\pageref{subsec:matched_ablation}\\[0.4em]
\makebox[1.5em][l]{}B.2. \textbf{Diffusion Denoising Steps} \dotfill \pageref{subsec:denoising_steps}\\[0.4em]
\makebox[1.5em][l]{}B.3. \textbf{VLT Augmentation} \dotfill \pageref{subsec:vlt_augmentation}\\[0.4em]
\makebox[1.5em][l]{}B.4. \textbf{Training Hyperparameters} \dotfill 
\pageref{subsec:training_hyperparameters}\\[0.4em]
\noindent C. \textbf{Additional Evaluations} \dotfill \pageref{sec:additional_evaluation}\\[0.4em]
\makebox[1.5em][l]{}C.1. \textbf{Video Frame Synthesis Baselines} \dotfill \pageref{subsec:video_frame_synthesis_baselines}\\[0.4em]
\makebox[1.5em][l]{}C.2. \textbf{Flow and Warping Error Analysis} \dotfill \pageref{subsec:flow_warping_analysis}\\[0.4em]
\makebox[1.5em][l]{}C.3. \textbf{Complexity Comparison} \dotfill \pageref{subsec:model_inference_comparison}\\[0.4em]
\makebox[1.5em][l]{}C.4. \textbf{Statistical Significance} \dotfill \pageref{subsec:statistical_significance}\\[0.4em]
\noindent D. \textbf{Reliability of Temporal Consistency Metrics} \dotfill 
\pageref{sec:temporal_consistency}\\[0.4em]

\section{Additional Implementation Details}
\label{sec:implementation_details}

\noindent\textbf{Training and Inference Setup.} 
We trained our model using 4 NVIDIA RTX A6000 GPUs, each equipped with 48 GB of memory.
Training was conducted in mixed precision (FP16) using the Hugging Face Accelerate framework for efficient memory usage.
During validation, the VAE decoder was temporarily upcast to FP32 for numerical stability, and loss computation was automatically performed in FP32 to prevent underflow.
We used gradient accumulation with 16 steps, with a per-GPU batch size of 1 resulting in a batch size of 64.
An Exponential Moving Average (EMA) with the default decay rate of 0.9999 was applied to the ControlNet parameters.
The Accelerate framework handled gradient synchronization across GPUs via NCCL all-reduce.
Four dataloader workers were assigned to each process (16 in total) to enable parallel data loading.
During inference, we adopted an iterative diffusion process with 5 denoising steps using an Euler sampler under the EDM formulation with a log-uniform $\sigma$ schedule from 3.0 to 0.05.
We set the guidance scale to 3.0 for classifier-free guidance and applied an anchor frame noise augmentation of 0.02. \\

\noindent\textbf{Event Representation.}
For each anchor-to-target interval, we construct a multi-stack event representation as
\[
\mathcal{F}_{\mathrm{event}} = \left\{
F^{(0)}, F^{(1)}, \ldots, F^{(M-1)}
\right\},
\]
where $F^{(0)}$ corresponds to the full-interval accumulation and $F^{(m)}$ captures increasingly short-term motion as $m$ increases.
In Figures~\ref{fig:overview} and~\ref{fig:training}, we use the terms \textit{Event Frame} and \textit{Event Frames} to distinguish these two usages.
\textit{Event Frames} refers to the multi-stack event representation $\mathcal{F}_{\mathrm{event}}$, which is provided as event conditioning to the diffusion model.
Among these stacks, the longest-window event frame $F^{(0)}$, accumulated over the entire anchor-to-target interval, is denoted as \textit{Event Frame}.
This full-interval event frame summarizes all motion and brightness changes between the anchor and target frames and is used by ER-VAE for event-to-residual alignment.
By contrast, the multi-stack \textit{Event Frames} provide motion cues at multiple temporal scales, allowing the event-conditioned diffusion model to capture both slow and fast motion patterns. \\

\noindent\textbf{Sampling Seed Robustness.}
We additionally analyze the effect of the initial Gaussian noise seed used during diffusion sampling under the 3-frame video frame prediction and interpolation settings.
The main experiments were performed with seed 123, and two additional runs with different noise seeds (0 and 42) were included for comparison.
As shown in Table~\ref{tab:sampling_seed_comparison}, variations in the initial noise seed led to only minor changes across all evaluation metrics, indicating that the model produces stable results under different diffusion sampling conditions.

\begin{table}[h]
\centering
\caption{Quantitative comparison across different diffusion sampling seeds on BS-ERGB~\cite{tulyakov2022timelenspp} under the 3-frame video frame prediction and interpolation settings.}
\label{tab:sampling_seed_comparison}
\renewcommand{\arraystretch}{1.2}
\resizebox{0.85\columnwidth}{!}{
\begin{tabular}{c|c|*{3}{>{\centering\arraybackslash}m{0.17\columnwidth}|}>{\centering\arraybackslash}m{0.17\columnwidth}}
\hline
Task & Metrics & Seed \#123 & Seed \#0 & Seed \#42 & Mean $\pm$ Std \\
\hline
\multirow{3}{*}{\makecell[c]{Video Frame\\Prediction}}
& PSNR $\uparrow$ & 24.20 & 23.92 & 24.09 & 24.07 $\pm$ 0.14 \\
& SSIM $\uparrow$ & 0.804 & 0.800 & 0.802 & 0.802 $\pm$ 0.002 \\
& LPIPS $\downarrow$ & 0.119 & 0.119 & 0.120 & 0.119 $\pm$ 0.001 \\
\hline
\multirow{3}{*}{\makecell[c]{Video Frame\\Interpolation}}
& PSNR $\uparrow$ & 24.28 & 24.23 & 24.28 & 24.26 $\pm$ 0.03 \\
& SSIM $\uparrow$ & 0.831 & 0.827 & 0.827 & 0.828 $\pm$ 0.002 \\
& LPIPS $\downarrow$ & 0.116 & 0.115 & 0.115 & 0.115 $\pm$ 0.001 \\
\hline
\end{tabular}
}
\renewcommand{\arraystretch}{1.0}
\end{table}
\section{Ablation Studies}
\label{sec:ablation_studies}

\subsection{Model Formulation and Event Conditioning}
\label{subsec:matched_ablation}
We further analyze the effects of residual modeling, event-informed initialization,
and event conditioning under matched model and training settings.
As shown in Table~\ref{tab:matched_ablation}, residual prediction consistently
outperforms direct target prediction under both Gaussian initialization
(D vs. A) and ER-aligned initialization (F vs. C).
Comparing D and F shows the benefit of ER-aligned initialization, while
E and F demonstrate that residual alignment provides more effective
initialization than generic VAE event features.
The comparisons D/G and F/H further show that explicit event conditioning
improves residual denoising under both Gaussian and event-informed
initialization.
Finally, F outperforms both the condition-only variant J and the
capacity-matched regressor I, supporting the combination of residual
prediction, ER-aligned initialization, and diffusion-based generation.

\begin{table}[t]
\centering
\caption{
Matched ablation study on BS-ERGB~\cite{tulyakov2022timelenspp} under the 3-frame video frame prediction setting,
analyzing residual modeling, initialization, event conditioning, and generator design.
}
\label{tab:matched_ablation}
\setlength{\tabcolsep}{2.5pt}
\renewcommand{\arraystretch}{1.0}
\resizebox{0.96\columnwidth}{!}{%
\begin{tabular}{cllclccc}
\hline
\noalign{\vskip 2pt}
ID & Target & Initialization & Event Cond. & Generator
& PSNR$\uparrow$ & SSIM$\uparrow$ & LPIPS$\downarrow$ \\
\noalign{\vskip 1pt}
\hline

A & $z_{t+1}$
  & $\mathcal{N}(0,I)$
  & \checkmark
  & Diffusion
  & 16.72 & 0.571 & 0.340 \\

B & $z_{t+1}$
  & $z_t+z_{\rm event}^{\rm VAE}$
  & \checkmark
  & Diffusion
  & 19.51 & 0.703 & 0.315 \\

C & $z_{t+1}$
  & $z_t+z_{\rm event}^{\rm ER}$
  & \checkmark
  & Diffusion
  & 19.96 & 0.711 & 0.313 \\

\hline

D & $z_{\rm res}$
  & $\mathcal{N}(0,I)$
  & \checkmark
  & Diffusion
  & 23.67 & 0.762 & 0.161 \\

E & $z_{\rm res}$
  & $z_{\rm event}^{\rm VAE}$
  & \checkmark
  & Diffusion
  & 21.43 & 0.734 & 0.204 \\

\textbf{F (\ours)}
  & $\mathbf{z_{\rm res}}$
  & $\mathbf{z_{\rm event}^{\rm ER}}$
  & \checkmark
  & \textbf{Diffusion}
  & \textbf{24.20}
  & \textbf{0.804}
  & \textbf{0.119} \\

G & $z_{\rm res}$
  & $\mathcal{N}(0,I)$
  & --
  & Diffusion
  & 23.27 & 0.760 & 0.156 \\

H & $z_{\rm res}$
  & $z_{\rm event}^{\rm ER}$
  & --
  & Diffusion
  & 18.96 & 0.704 & 0.236 \\

I & $z_{\rm res}$
  & $z_{\rm event}^{\rm ER}$
  & \checkmark
  & Regressor
  & 21.01 & 0.749 & 0.152 \\

J & $z_{\rm res}$
  & $\mathcal{N}(0,I)$
  & $z_{\rm event}^{\rm ER}$
  & Diffusion
  & 22.93 & 0.741 & 0.169 \\

\hline
\end{tabular}%
}
\vspace{-2pt}
\end{table}

\subsection{Diffusion Denoising Steps}
\label{subsec:denoising_steps}
We analyze the effect of the number of diffusion denoising steps in Table~\ref{tab:denoising_steps}.
All results are measured on BS-ERGB~\cite{tulyakov2022timelenspp} under the 3-frame video frame prediction setting using a single NVIDIA RTX 3090 GPU at a resolution of $970 \times 625$.
We evaluate denoising steps of 2, 5, 10, 15, 25, and 50 to analyze the quality-efficiency trade-off.
The 5-step setting achieves the best overall balance, improving PSNR and SSIM over the 2-step setting while maintaining comparable LPIPS.
Using fewer steps slightly reduces PSNR and SSIM, while using more steps brings negligible quality gains but substantially increases inference time.
Therefore, we use 5 denoising steps as the default setting.

\begin{table}[h]
\centering
\caption{Ablation on the number of diffusion denoising steps on BS-ERGB~\cite{tulyakov2022timelenspp} under the 3-frame video frame prediction setting. The current setting uses 5 denoising steps.}
\label{tab:denoising_steps}
\renewcommand{\arraystretch}{1.2}
\resizebox{0.75\columnwidth}{!}{
\begin{tabular}{c|*{5}{>{\centering\arraybackslash}m{0.12\columnwidth}|}>{\centering\arraybackslash}m{0.12\columnwidth}}
\hline
Steps & 2 & 5 (default) & 10 & 15 & 25 & 50 \\
\hline
PSNR $\uparrow$ & 24.15 & 24.20 & 24.20 & \textbf{24.21} & \textbf{24.21} & \textbf{24.21} \\
SSIM $\uparrow$ & 0.800 & \textbf{0.804} & 0.802 & 0.801 & 0.801 & 0.801 \\
LPIPS $\downarrow$ & \textbf{0.119} & \textbf{0.119} & 0.120 & 0.120 & 0.120 & 0.120 \\
Time (s) & 7.1 & 10.3 & 17.7 & 22.9 & 35.4 & 67.5 \\
\hline
\end{tabular}
}
\renewcommand{\arraystretch}{1.0}
\end{table}
\vspace{-1.5em}

\subsection{VLT Augmentation}
\label{subsec:vlt_augmentation}
\noindent\textbf{Temporal Processing Order.} We first investigate the impact of the interval sequencing during training, as summarized in Table~\ref{tab:vlt_order_a}. 
For this analysis, the three training phases follow a fixed 5:3:2 step ratio according to the listed order.
The $1 \to 2 \to 3$ order achieves the best overall performance, supporting the curriculum-style design that gradually increases the temporal interval from short to long.
The reversed order, $3 \to 2 \to 1$, generally achieves the second-best performance, while the non-consecutive order, $1 \to 3 \to 2$, performs worse.
This suggests that a smooth progression of temporal difficulty is beneficial for stable optimization. 

\noindent\textbf{Ratio Configuration.} Using the best-performing $1 \to 2 \to 3$ order, we further evaluate different training-step allocation ratios in Table~\ref{tab:vlt_ratio_b}.
The 5:3:2 ratio provides the best overall balance across 1-frame and 3-frame prediction settings.
Compared with the uniform 1:1:1 ratio and the long-interval-heavy 2:3:5 ratio, allocating more training steps to shorter intervals provides more stable reconstruction quality.
These results show that VLT benefits from allocating more training steps to shorter intervals before gradually increasing the temporal range.

\begin{table}[t]
    \centering
    \caption{Ablation study of VLT Augmentation on BS-ERGB~\cite{tulyakov2022timelenspp}. Results are reported for 1-frame / 3-frame video frame prediction. The best and second-best results are highlighted in \textbf{bold} and \underline{underlined}, respectively.}
    \label{tab:vlt_order}
    \label{tab:vlt_ratio}
    \makeatletter
    \protected@edef\@currentlabel{\thetable-(a)}\label{tab:vlt_order_a}
    \protected@edef\@currentlabel{\thetable-(b)}\label{tab:vlt_ratio_b}
    \makeatother
    \renewcommand{\arraystretch}{1.2}
    \scriptsize
    \noindent\textbf{(a) Temporal processing order.}\vspace{0.5em}

    \resizebox{0.85\columnwidth}{!}{%
    \begin{tabular}{>{\centering\arraybackslash}m{0.16\columnwidth}|>{\centering\arraybackslash}m{0.16\columnwidth}|>{\centering\arraybackslash}m{0.16\columnwidth}|>{\centering\arraybackslash}m{0.16\columnwidth}}
    \hline
    Order & PSNR $\uparrow$ & SSIM $\uparrow$ & LPIPS $\downarrow$ \\
    \hline
    $1 \rightarrow 2 \rightarrow 3$ & \textbf{25.27} / \textbf{24.20} & \textbf{0.895} / \textbf{0.804} & \textbf{0.100} / \textbf{0.119} \\
    \hline
    $3 \rightarrow 2 \rightarrow 1$ & \underline{22.58} / \underline{22.07} & \underline{0.775} / \underline{0.735} & \underline{0.150} / 0.181 \\
    \hline
    $1 \rightarrow 3 \rightarrow 2$ & 21.39 / 20.79 & 0.755 / 0.701 & 0.159 / \underline{0.179} \\
    \hline
    \end{tabular}
    }

    \vspace{1.5em}
    \noindent\textbf{(b) Ratio configuration.}\vspace{0.5em}

    \resizebox{0.85\columnwidth}{!}{%
    \begin{tabular}{>{\centering\arraybackslash}m{0.16\columnwidth}|>{\centering\arraybackslash}m{0.16\columnwidth}|>{\centering\arraybackslash}m{0.16\columnwidth}|>{\centering\arraybackslash}m{0.16\columnwidth}}
    \hline
    Ratio & PSNR $\uparrow$ & SSIM $\uparrow$ & LPIPS $\downarrow$ \\
    \hline
    5:3:2 & \textbf{25.27} / \textbf{24.20} & \textbf{0.895} / \textbf{0.804} & \textbf{0.100} / \textbf{0.119} \\
    \hline
    1:1:1 & \underline{24.13} / 23.00 & 0.752 / 0.734 & 0.140 / 0.164 \\
    \hline
    2:3:5 & 23.95 / \underline{23.05} & \underline{0.785} / \underline{0.745} & \underline{0.139} / \underline{0.163} \\
    \hline
    \end{tabular}
    }
    \normalsize
    \renewcommand{\arraystretch}{1.0}
\end{table}

\subsection{Training Hyperparameters}
\label{subsec:training_hyperparameters}

\noindent\textbf{Stage~1: Event-to-Residual Alignment VAE (ER-VAE).}
We provide quantitative analysis of its training objective hyperparameters in Table~\ref{tab:ablation_vae}.
We compare ER-VAE variants under a fixed stage~2 diffusion model to isolate the effect of stage~1 hyperparameters.
We evaluate three design choices: latent-space scaling, the use of KL divergence, and the image reconstruction loss weight $\gamma$. 

First, applying a latent-space scaling factor to the ER-VAE output improves reconstruction quality by making the aligned event latent more compatible with the pretrained VAE latent space.
Without this scaling, performance decreases across both 1-frame and 3-frame settings. 

Second, removing the KL divergence term leads to better reconstruction quality, supporting our deterministic event-to-residual mapping.
In this task, KL regularization can act as an unnecessary bottleneck that constrains the event representation before residual alignment. 

Finally, we evaluate fixed values of the image loss weight $\gamma$, which controls the balance between latent alignment and pixel-level reconstruction.
The results show that $\gamma$ values from 0.3 to 0.7 achieve consistently strong performance, with only minor differences across metrics.
Based on this observation, we adopt a cosine schedule that gradually increases $\gamma$ from 0.3 to 0.7 during training, allowing the model to first emphasize latent alignment and then progressively strengthen pixel-level reconstruction.

\begin{table}[t]
    \centering
    \caption{Ablation on VAE scaling, KL divergence objective, and image loss weight for BS-ERGB~\cite{tulyakov2022timelenspp}. Metrics are reported for 1-frame and 3-frame prediction tasks. The best results are highlighted in \textbf{bold}.}
    \label{tab:ablation_vae}
    \renewcommand{\arraystretch}{1.2}
    \resizebox{0.62\columnwidth}{!}{
    \begin{tabular}{c|ccc|ccc}
    \hline
    \multirow{2}{*}{Setting} & \multicolumn{3}{c|}{1 frame} & \multicolumn{3}{c}{3 frames} \\
\cline{2-7}
    & PSNR$\uparrow$ & SSIM$\uparrow$ & LPIPS$\downarrow$ 
    & PSNR$\uparrow$ & SSIM$\uparrow$ & LPIPS$\downarrow$ \\
    \hline
    \multicolumn{7}{c}{VAE Scaling} \\
    \hline
     w/o & 25.20 & 0.882 & \textbf{0.100} & 24.13 & 0.801 & \textbf{0.119} \\
     with & \textbf{25.27} & \textbf{0.895} & \textbf{0.100} & \textbf{24.20} & \textbf{0.804} & \textbf{0.119} \\
    
    \hline
    \multicolumn{7}{c}{KL Divergence} \\
    \hline
    w/o & \textbf{25.27} & \textbf{0.895} & \textbf{0.100} & \textbf{24.20} & \textbf{0.804} & \textbf{0.119} \\
    with & 25.11 & 0.892 & 0.104 & 24.16 & \textbf{0.804} & \textbf{0.119} \\
    
    \hline
    \multicolumn{7}{c}{Image Loss Weight} \\
    \hline
    0.1 & 25.26 & 0.881 & 0.101 & 24.08 & 0.800 & 0.121 \\
    0.3 & \textbf{25.27} & \textbf{0.895} & \textbf{0.100} & \textbf{24.20} & \textbf{0.804} & \textbf{0.119} \\
    0.5 & \textbf{25.27} & 0.893 & 0.101 & \textbf{24.20} & \textbf{0.804} & \textbf{0.119} \\
    0.7 & 25.26 & 0.893 & \textbf{0.100} & 24.19 & 0.801 & \textbf{0.119} \\
    \hline
    \end{tabular}
    }
\end{table}

\noindent\textbf{Stage~2: Event-Conditional Diffusion Model.}
\label{subsec:lpips_weight}
We further analyze the LPIPS loss weight $\lambda$ used in the stage~2 diffusion objective.
This ablation is performed after the initial convergence of stage~1, using the same ER-VAE setting while varying only $\lambda$ in the diffusion training objective.
As shown in Table~\ref{tab:lpips_weight}, $\lambda=0.3$ achieves the best SSIM and LPIPS, while also maintaining competitive PSNR.
Although $\lambda=0.5$ gives a slightly higher PSNR, its perceptual quality is worse than $\lambda=0.3$.
Therefore, we use $\lambda=0.3$ as the default setting, which provides the best overall trade-off between reconstruction fidelity and perceptual quality.

\begin{table}[h]
\centering
\caption{Ablation on the LPIPS loss weight $\lambda$ in stage~2 training. The ablation is performed on BS-ERGB~\cite{tulyakov2022timelenspp} under the 3-frame video frame prediction setting. The best results are highlighted in \textbf{bold}.}
\label{tab:lpips_weight}
\renewcommand{\arraystretch}{1.2}
\resizebox{0.57\columnwidth}{!}{
\begin{tabular}{c|c|c|c|c|c}
\hline
Metric & $\lambda=0.1$ & $\lambda=0.3$ & $\lambda=0.5$ & $\lambda=0.7$ & $\lambda=0.9$ \\
\hline
PSNR$\uparrow$ & 22.51 & 23.01 & \textbf{23.02} & 22.47 & 22.42 \\
SSIM$\uparrow$ & 0.725 & \textbf{0.746} & 0.738 & 0.735 & 0.698 \\
LPIPS$\downarrow$ & 0.182 & \textbf{0.160} & 0.169 & 0.168 & 0.178 \\
\hline
\end{tabular}
}
\renewcommand{\arraystretch}{1.0}
\end{table}

\section{Additional Evaluations}
\label{sec:additional_evaluation}

\subsection{Video Frame Synthesis Baselines}
\label{subsec:video_frame_synthesis_baselines}
In Section~\ref{sec:experiments}, we compare \ours\ with representative task-specific baselines for video frame prediction and interpolation.
Here, we provide broader quantitative comparisons with methods using different input modalities and task formulations, including RGB-only frame prediction, event-only frame reconstruction, and event-based optical-flow-and-warping approaches.
Since these methods differ in their input settings and synthesis formulations, the results should be interpreted as additional reference comparisons.
Tables~\ref{tab:supp_results_real} and~\ref{tab:supp_results_gopro} report the results on the real and synthetic event datasets, respectively.
The results highlight the complementary roles of RGB frames and event observations.

\begin{table*}[b]
\centering
\caption{Quantitative comparison for video frame prediction on the real event datasets BS-ERGB~\cite{tulyakov2022timelenspp} and HS-ERGB~\cite{tulyakov2021time}.}
\label{tab:supp_results_real}
\small
\renewcommand{\arraystretch}{1.15}
\resizebox{\textwidth}{!}{
\begin{tabular}{c|c|c|ccc|ccc|ccc}
\hline
\multirow{3}{*}{Modality} & \multirow{3}{*}{Task} & \multirow{3}{*}{Method} &
\multicolumn{6}{c|}{BS-ERGB~\cite{tulyakov2022timelenspp}} &
\multicolumn{3}{c}{HS-ERGB~\cite{tulyakov2021time}} \\
\cline{4-12}
& & & \multicolumn{3}{c|}{1 frame} & \multicolumn{3}{c|}{3 frames} &
\multicolumn{3}{c}{7 frames} \\
\cline{4-12}
& & & PSNR$\uparrow$ & SSIM$\uparrow$ & LPIPS$\downarrow$ &
PSNR$\uparrow$ & SSIM$\uparrow$ & LPIPS$\downarrow$ &
PSNR$\uparrow$ & SSIM$\uparrow$ & LPIPS$\downarrow$ \\
\hline
\multirow{2}{*}{2 Images}
& \multirow{2}{*}{\shortstack{Frame\\Prediction}}
& RaMViD~\cite{hoppe2022ramvid} (TMLR'22)
& 14.40 & 0.451 & 0.400 & 14.32 & 0.450 & 0.392 & 19.28 & 0.691 & 0.293 \\
& & DMVFN~\cite{hu2023dmvfn} (CVPR'23)
& 23.67 & 0.675 & 0.093 & 22.08 & 0.652 & 0.125 & 28.14 & 0.788 & 0.122 \\
\hline
\multirow{2}{*}{Event}
& \multirow{2}{*}{Reconstruction}
& E2HQV~\cite{qu2024e2hqv} (AAAI'24)
& 14.19 & 0.358 & 0.247 & 13.88 & 0.358 & 0.260 & 17.91 & 0.487 & 0.329 \\
& & HyperE2VID~\cite{ercan2024hypere2vid} (TIP'24)
& 18.79 & 0.487 & 0.273 & 18.78 & 0.410 & 0.278 & 18.20 & 0.529 & 0.331 \\
\hline
\multirow{2}{*}{\shortstack{1 Image\\+ Event}}
& \multirow{2}{*}{\shortstack{Optical Flow\\Estimation}}
& DCEIFlow~\cite{wan2022learning} (TIP'22)
& 20.29 & 0.715 & 0.132 & 18.98 & 0.678 & 0.175 & 23.40 & 0.746 & 0.174 \\
& & bFlow~\cite{gehrig2024dense} (TPAMI'24)
& 20.56 & 0.648 & 0.121 & 19.37 & 0.673 & 0.150 & 24.15 & 0.682 & 0.161 \\
\hline
\end{tabular}
}
\renewcommand{\arraystretch}{1.0}
\end{table*}

These comparisons also provide insight into the role of each input modality in video frame synthesis.
RGB-only methods exploit appearance cues such as color, texture, and scene layout, but lack direct high-temporal-resolution motion information between frames.
Event-only reconstruction methods capture fine-grained temporal changes, but recovering full RGB appearance from events alone remains challenging because events primarily encode brightness changes.
Event-based optical-flow-and-warping approaches exploit event motion cues, but their performance is still constrained by the accuracy of the estimated flow and the limitations of warping under occlusion and large motion.
In contrast, \ours\ combines RGB appearance information with event-based motion cues through inter-frame residual alignment, enabling more accurate synthesis across different datasets and temporal intervals.

\begin{table}[t]
\centering
\caption{Quantitative comparison for video frame prediction on the synthetic event dataset GoPro~\cite{nah2017deep}.}
\label{tab:supp_results_gopro}
\small
\renewcommand{\arraystretch}{1.15}
\resizebox{0.95\columnwidth}{!}{
\begin{tabular}{c|c|c|ccc|ccc}
\hline
\multirow{3}{*}{Modality} & \multirow{3}{*}{Task} & \multirow{3}{*}{Method} &
\multicolumn{6}{c}{GoPro~\cite{nah2017deep}} \\
\cline{4-9}
& & & \multicolumn{3}{c|}{7 frames} & \multicolumn{3}{c}{15 frames} \\
\cline{4-9}
& & & PSNR$\uparrow$ & SSIM$\uparrow$ & LPIPS$\downarrow$ &
PSNR$\uparrow$ & SSIM$\uparrow$ & LPIPS$\downarrow$ \\
\hline
\multirow{2}{*}{2 Images}
& \multirow{2}{*}{\shortstack{Frame\\Prediction}}
& RaMViD~\cite{hoppe2022ramvid} (TMLR'22)
& 13.50 & 0.487 & 0.457 & 13.32 & 0.484 & 0.560 \\
& & DMVFN~\cite{hu2023dmvfn} (CVPR'23)
& 14.65 & 0.442 & 0.451 & 13.12 & 0.413 & 0.559 \\
\hline
\multirow{2}{*}{Event}
& \multirow{2}{*}{Reconstruction}
& E2HQV~\cite{qu2024e2hqv} (AAAI'24)
& 10.81 & 0.468 & 0.423 & 10.86 & 0.471 & 0.520 \\
& & HyperE2VID~\cite{ercan2024hypere2vid} (TIP'24)
& 9.72 & 0.430 & 0.468 & 9.62 & 0.430 & 0.571 \\
\hline
\multirow{2}{*}{\shortstack{1 Image\\+ Event}}
& \multirow{2}{*}{\shortstack{Optical Flow\\Estimation}}
& DCEIFlow~\cite{wan2022learning} (TIP'22)
& 12.26 & 0.385 & 0.547 & 11.83 & 0.373 & 0.569 \\
& & bFlow~\cite{gehrig2024dense} (TPAMI'24)
& 12.82 & 0.527 & 0.508 & 12.31 & 0.641 & 0.543 \\
\hline
\end{tabular}
}
\renewcommand{\arraystretch}{1.0}
\end{table}

\subsection{Flow and Warping Error Analysis}
\label{subsec:flow_warping_analysis}
To further analyze the limitations of flow-based event-guided synthesis,
we decompose VFPSIE~\cite{zhu2024video} into its flow-warping and refinement stages under different motion and occlusion conditions.
As a diagnostic reference, we estimate optical flow with RAFT using the
anchor and ground-truth target frames and apply the same warping operator
as VFPSIE.
As shown in Table~\ref{tab:flow_diagnosis}, the gap between VFPSIE-predicted flow and the RAFT flow increases from 1.02 dB for small motion to 4.08 dB for large motion, indicating that flow estimation becomes increasingly difficult as motion grows.
Although the refinement stage partially recovers the warping errors,
its performance remains limited in large-motion and occluded regions.
In contrast, \ours\ improves PSNR over the final VFPSIE output by
4.95 dB and 3.83 dB on large-motion and occluded regions, respectively.

\begin{table}[t]
\centering
\caption{
Flow and warping error analysis of VFPSIE~\cite{zhu2024video}
on HS-ERGB using 1,500 samples (PSNR $\uparrow$).
For the reference flow, RAFT is applied to the anchor and ground-truth
target frames, followed by the same warping operator as VFPSIE.
}
\label{tab:flow_diagnosis}
\setlength{\tabcolsep}{2pt}
\renewcommand{\arraystretch}{1.0}
\resizebox{0.66\columnwidth}{!}{%
\begin{tabular}{ccccc}
\hline
\noalign{\vskip 3pt}
Stage & Small & Medium & Large & Occlusion \\
\noalign{\vskip 1.5pt}
\hline
VFPSIE Pred. Flow Warp
    & 30.24 & 27.54 & 20.95 & 23.63 \\
RAFT Flow Warp
    & 31.26 & 29.98 & 25.03 & 25.01 \\
VFPSIE Final
    & 30.55 & 28.47 & 22.68 & 25.34 \\
\textbf{\ours}
    & \textbf{33.87} & \textbf{31.85} & \textbf{27.63} & \textbf{29.17} \\
\hline
\end{tabular}
}
\end{table}

\subsection{Complexity Comparison}
\label{subsec:model_inference_comparison}
Table~\ref{tab:model_inference_comparison} compares the computational complexity and inference efficiency of representative frame prediction and interpolation methods.
All models were evaluated on a single NVIDIA RTX 3090 GPU by generating a $970 \times 625$ resolution frame on BS-ERGB~\cite{tulyakov2022timelenspp}, with results averaged over five independent runs.
The comparison reveals a distinct trade-off between traditional optical flow-based methods and recent diffusion-based approaches. 
Optical flow methods, such as TimeLens~\cite{tulyakov2021time}, CBMNet~\cite{kim2023cbmnet}, and VFPSIE~\cite{zhu2024video} generally exhibit lower latency and memory footprints. 
In contrast, diffusion-based methods require higher computation because of iterative denoising.
Nevertheless, our model remains substantially more efficient than the video diffusion baseline RE-VDM~\cite{chen2025repurposing}.
For frame interpolation, our model reduces the parameter count by 30.9\%, GFLOPs by 12.4\%, memory usage by 54.1\%, and inference time by approximately $3.83\times$ compared with RE-VDM.
Moreover, despite being a diffusion-based model, our model uses less memory than CBMNet-Large, indicating that our residual-latent formulation and 5-step inference setting provide a favorable trade-off between synthesis quality and efficiency.

\begin{table*}[t]
    \centering
    \caption{Model GFLOPs, parameter count, run time, and memory usage comparison.}
    \label{tab:model_inference_comparison}
    \renewcommand{\arraystretch}{1.4}
    \resizebox{0.95\textwidth}{!}{
    \begin{tabular}{c|c|c|c|c|c|c}
    \hline
    Task & Method & GFLOPs & \# Params (M) & Run Time (s) & \makecell{Memory \\ Usage (GB)} & Notes \\
    \hline
    \multirow{2}{*}{\makecell{Frame Prediction}}
    & VFPSIE~\cite{zhu2024video} & 55.3 & 2.15 & 0.0147 & 0.457 & Optical flow-based \\
    \cline{2-7}
    & \ours \textbf{(VFP)} & 10275.0 & 2029.76 & 10.3 & 4.310 & Diffusion Model \\
    \hline
    \multirow{4}{*}{\makecell{Frame Interpolation}}
    & TimeLens~\cite{tulyakov2021time} & 2456.4 & 79.21 & 0.133 & 1.73 & Optical flow-based \\
    \cline{2-7}
    & CBMNet-Large~\cite{kim2023cbmnet} & 3201.1 & 22.23 & 1.163 & 10.161 & Optical flow-based \\
    \cline{2-7}
    & RE-VDM~\cite{chen2025repurposing} & 23921.9 & 2936.70 & 82.296 & 9.388 & Video Diffusion Model \\
    \cline{2-7}
    & \ours \textbf{(VFI)} & 20953.1 & 2029.76 & 21.46 & 4.310 & Diffusion Model \\
    \hline
    \end{tabular}
    }
    \renewcommand{\arraystretch}{1.0}
    \vspace{-1em}
\end{table*}

\subsection{Statistical Significance}
\label{subsec:statistical_significance}
We further evaluate the statistical significance of the performance gains using sequence-level metrics.
For each sequence, we first compute the average PSNR, SSIM, and LPIPS over all evaluated frames, and then perform statistical tests across sequences.
For PSNR and SSIM, mean improvement is computed as the increase over the baseline, while for LPIPS it is computed as the reduction from the baseline.
For video frame prediction, we compare our model against VFPSIE~\cite{zhu2024video} under the 7-frame setting.
As shown in Table~\ref{tab:stat_vfp_a}, our model achieves consistent sequence-level improvements, with statistically significant gains under both the Wilcoxon signed-rank test and paired $t$-test ($p < 0.005$ for all metrics).
For video frame interpolation, we compare our model against CBMNet-Large~\cite{kim2023cbmnet}, the strongest competing interpolation baseline on HS-ERGB, under the 7-frame setting.
As shown in Table~\ref{tab:stat_vfi_b}, the improvements are also statistically significant under both tests ($p < 0.005$ for all metrics), confirming that the observed gains are consistent at the sequence level rather than being caused by a small number of examples.

\begin{table}[h]
    \centering
    \caption{Sequence-level statistical significance tests on HS-ERGB~\cite{tulyakov2021time}. We report win counts, mean improvements, Wilcoxon signed-rank test $p$-values, and paired $t$-test $p$-values over 15 test sequences.}
    \label{tab:stat_vfp}
    \label{tab:stat_vfi}
    \makeatletter
    \protected@edef\@currentlabel{\thetable-(a)}\label{tab:stat_vfp_a}
    \protected@edef\@currentlabel{\thetable-(b)}\label{tab:stat_vfi_b}
    \makeatother
    \renewcommand{\arraystretch}{1.2}
    \scriptsize
    \noindent\textbf{(a) Video frame prediction against VFPSIE~\cite{zhu2024video} on the 7-frame setting.}\vspace{0.5em}

    \resizebox{0.72\columnwidth}{!}{
    \begin{tabular}{c|c|c|c|c}
    \hline
    Metric & Win Count & Mean Improvement & Wilcoxon $p$ & $t$-test $p$ \\
    \hline
    PSNR$\uparrow$ & 15 / 15 & 0.930 & 0.00051 & 0.00238 \\
    SSIM$\uparrow$ & 13 / 15 & 0.014 & 0.00098 & 0.00047 \\
    LPIPS$\downarrow$ & 13 / 15 & 0.011 & 0.00092 & 0.00044 \\
    \hline
    \end{tabular}
    }

    \vspace{1em}
    \noindent\textbf{(b) Video frame interpolation against CBMNet-Large~\cite{kim2023cbmnet} on the 7-frame setting.}\vspace{0.5em}

    \resizebox{0.72\columnwidth}{!}{
    \begin{tabular}{c|c|c|c|c}
    \hline
    Metric & Win Count & Mean Improvement & Wilcoxon $p$ & $t$-test $p$ \\
    \hline
    PSNR$\uparrow$ & 15 / 15 & 1.740 & 0.00033 & 0.00177 \\
    SSIM$\uparrow$ & 15 / 15 & 0.017 & 0.00029 & 0.00138 \\
    LPIPS$\downarrow$ & 15 / 15 & 0.052 & 0.00027 & 0.00086 \\
    \hline
    \end{tabular}
    }
    \renewcommand{\arraystretch}{1.0}
\end{table}
\vspace{-1em}

\section{Reliability of Temporal Consistency Metrics}
\label{sec:temporal_consistency}
As discussed in Section~\ref{subsec:results}, we evaluated temporal consistency using the discrepancy between predicted and ground-truth optical flow. 
This metric was adopted because commonly used video consistency metrics can behave unreliably in the context of video frame prediction.
In many video generation studies, temporal consistency is evaluated using metrics such as VBench~\cite{huang2023vbench}.
However, we observe that these metrics may assign higher consistency scores to predicted sequences than to the corresponding ground-truth sequences. 

\begin{figure}[h]
    \centering
    \begin{subfigure}{0.32\columnwidth}
        \centering
        \includegraphics[width=\linewidth]{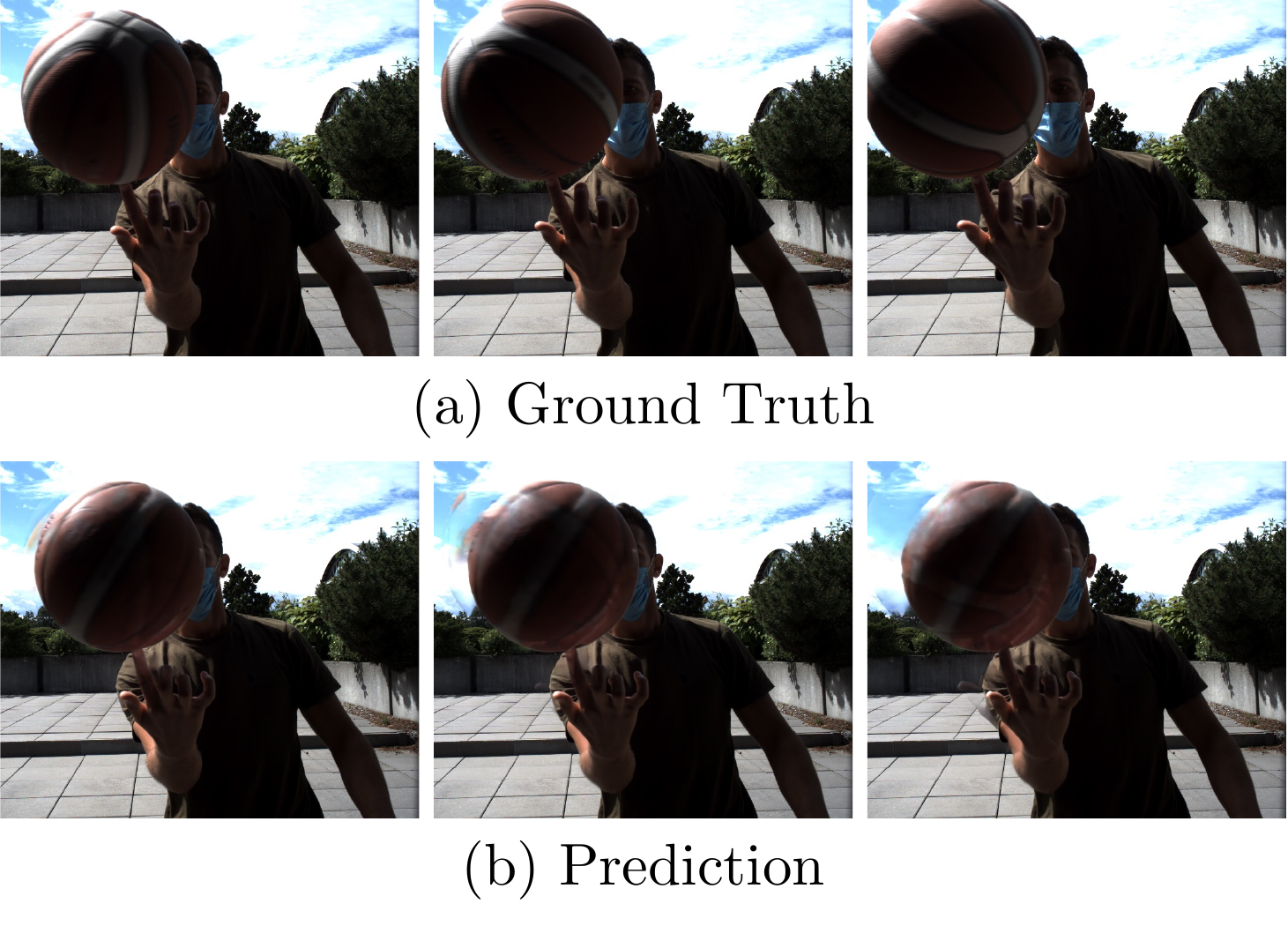}
        \caption{Basketball example.}
        \label{fig:vbench_ex1}
    \end{subfigure}
    \hspace{0.02\columnwidth}
    \begin{subfigure}{0.32\columnwidth}
        \centering
        \includegraphics[width=\linewidth]{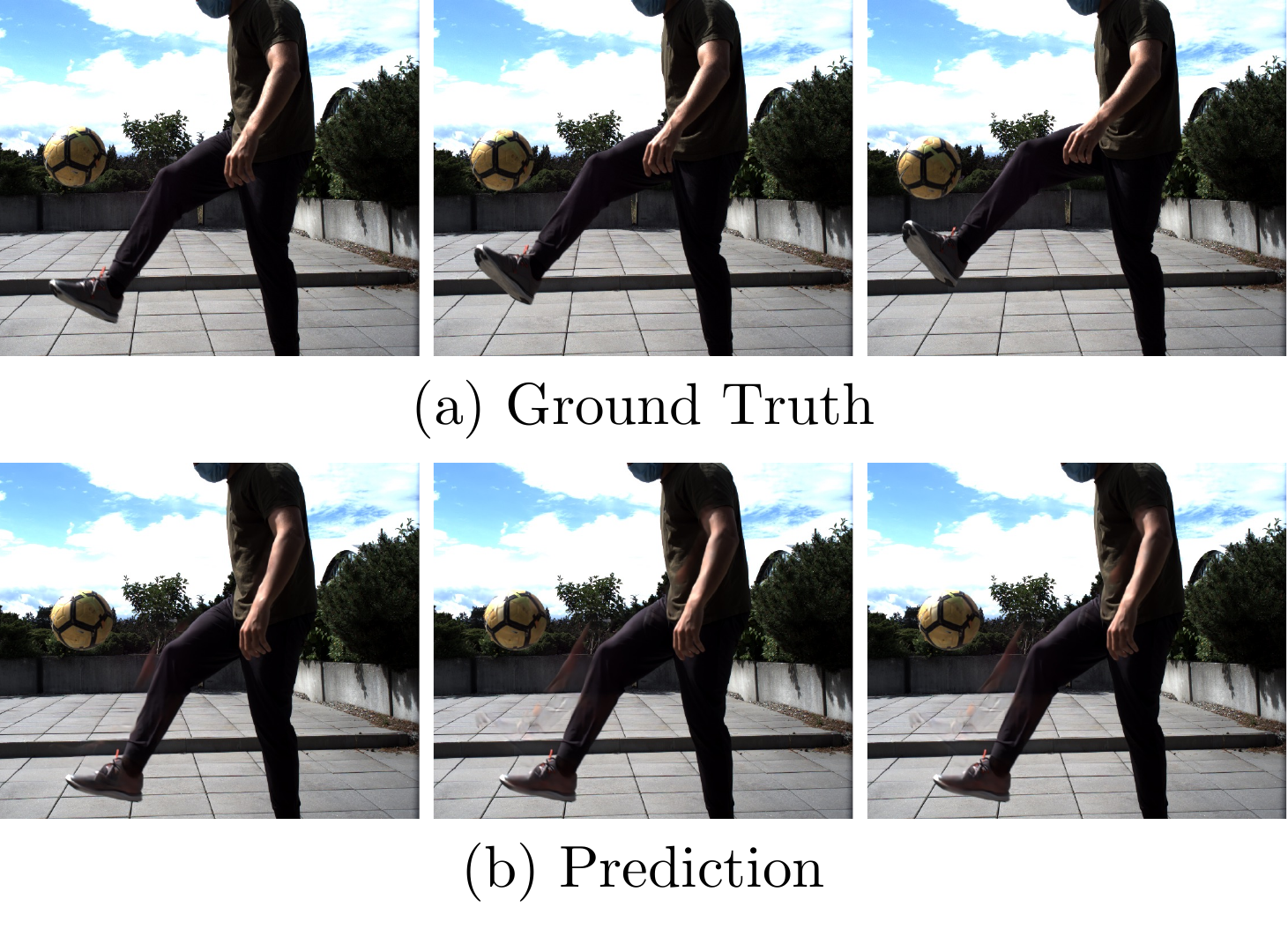}
        \caption{Football example.}
        \label{fig:vbench_ex2}
    \end{subfigure}
    \caption{Qualitative examples of ground-truth and predicted frame sequences.}
    \label{fig:vbench_examples}
\end{figure}

To illustrate this issue, we present two examples in 
Figure~\ref{fig:vbench_examples} with the corresponding VBench scores summarized in Table~\ref{tab:vbench_results}. 
As shown in the figure, the predicted sequences appear more static with reduced motion variation compared to the ground-truth sequences. 
Because VBench evaluates temporal consistency based on similarity between consecutive frames, such reduced motion can artificially increase the consistency score. 
For this reason, our evaluation instead measures the discrepancy between predicted and ground-truth optical flow fields, which directly reflects how accurately the motion dynamics are reproduced.

\vspace{-2.5em}

\begin{table}[h]
\centering
\caption{Examples where VBench~\cite{huang2023vbench} assigns higher temporal consistency scores to predicted sequences than to ground-truth sequences.}
\label{tab:vbench_results}
\renewcommand{\arraystretch}{1.2}
\resizebox{0.37\columnwidth}{!}{
\begin{tabular}{c|c|c}
\hline
Example & Ground Truth & Prediction \\
\hline
Basketball & 0.968 & 0.994 \\
Football & 0.965 & 0.996 \\
\hline
\end{tabular}}
\end{table}

\end{document}